\documentclass[10pt]{article} 
\usepackage[preprint]{tmlr}

\usepackage{amsmath,amsfonts,bm}

\def\eqref#1{equation~\ref{#1}}

\def\1{\bm{1}}

\DeclareMathAlphabet{\mathsfit}{\encodingdefault}{\sfdefault}{m}{sl}
\SetMathAlphabet{\mathsfit}{bold}{\encodingdefault}{\sfdefault}{bx}{n}

\usepackage{hyperref}
\usepackage{url}
\usepackage{graphicx}
\usepackage{adjustbox}
\usepackage[utf8]{inputenc}
\usepackage[T2A,T5,T1]{fontenc} 
\usepackage[russian,vietnamese,english]{babel}
\usepackage{CJKutf8}
\usepackage{kotex}   

\title{Equivalent Linear Mappings of Large Language Models}

\author{\name James R. Golden \email jamesgolden1@gmail.com \\
      \addr Oakland, CA}

\newcommand{\vect}[1]{\mathbf{#1}}

\def\month{10}  
\def\year{2025} 
\def\openreview{\url{https://openreview.net/forum?id=oDWbJsIuEp}} 

\begin{document}

\maketitle

\begin{abstract}
Despite significant progress in transformer interpretability, an understanding of the computational mechanisms of large language models (LLMs) remains a fundamental challenge. Many approaches interpret a network's hidden representations but remain agnostic about how those representations are generated. We address this by mapping LLM inference for a given input sequence to an equivalent and interpretable linear system which reconstructs the predicted output embedding with relative error below $10^{-13}$ at double floating-point precision, requiring no additional model training. We exploit a property of transformer decoders wherein every operation (gated activations, attention, and normalization) can be expressed as $A(x) \cdot x$, where $A(x)$ represents an input-dependent linear transform and $x$ preserves the linear pathway. To expose this linear structure, we strategically detach components of the gradient computation with respect to an input sequence, freezing the $A(x)$ terms at their values computed during inference, such that the Jacobian yields an equivalent linear mapping. This ``detached’’ Jacobian of the model reconstructs the output with one linear operator per input token, which is shown for Qwen 3, Gemma 3 and Llama 3, up to Qwen 3 14B. These linear representations demonstrate that LLMs operate in extremely low-dimensional subspaces where the singular vectors can be decoded to interpretable semantic concepts. The computation for each intermediate output also has a linear equivalent, and we examine how the linear representations of individual layers and their attention and multilayer perceptron modules build predictions, and use these as steering operators to insert semantic concepts into unrelated text. Despite their expressive power and global nonlinearity, modern LLMs can be interpreted through equivalent linear representations that reveal low-dimensional semantic structures in the next-token prediction process. Code is available at \url{https://github.com/jamesgolden1/equivalent-linear-LLMs/} .
\end{abstract}\section{Introduction}

\begin{figure}[!t]
\centering
\includegraphics[width=0.9\columnwidth]{"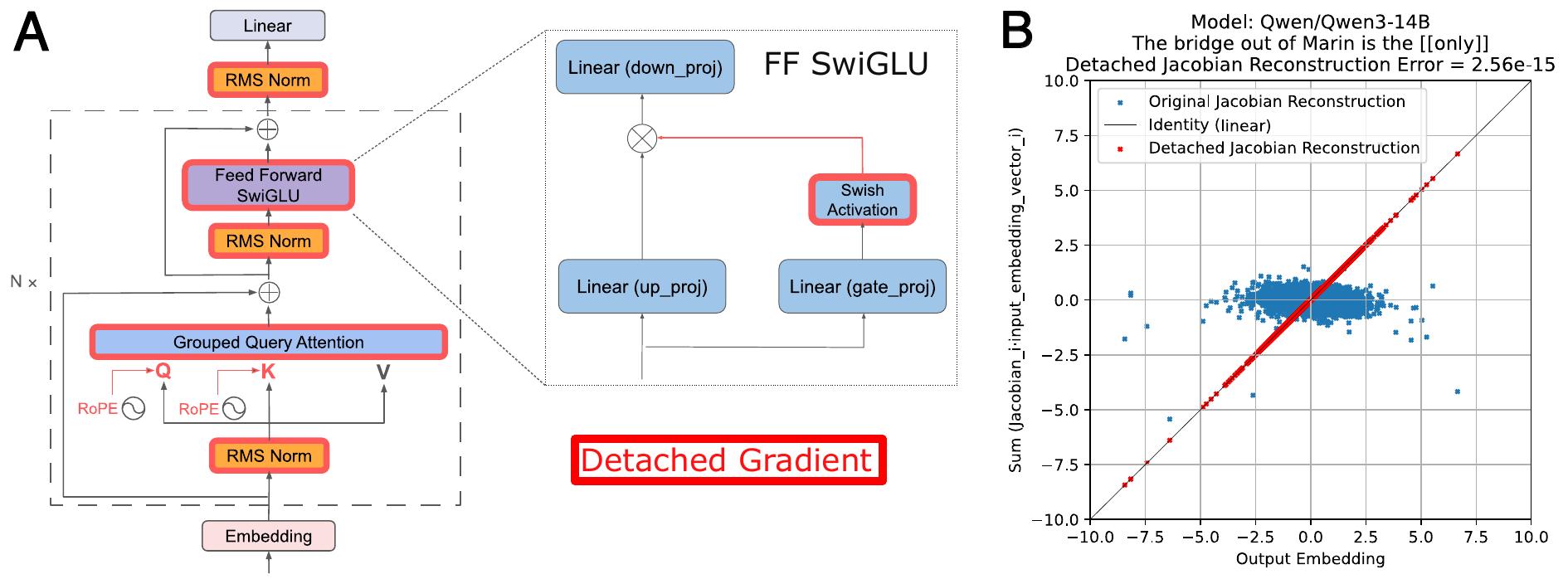"}
\caption{A) A schematic of the transformer decoder \citep{grattafiori2024llama, nvidia_te_llama_tutorial}. The PyTorch gradient detach operations for components outlined in red effectively freeze the nonlinear activations for a given input sequence, creating a linear path for the gradient with respect to the input embedding vectors, but do not change the output. The output embedding prediction can be mapped to an equivalent linear system by the Jacobian autograd operation. The feedforward module with a gated linear activation function is shown in expanded form to demonstrate how the gating term can be detached from the gradient to form a linear path, achieving linearity for a given input. The RMSNorm layers and softmax attention blocks also must be detached from the gradient. B) For the input sequence ``The bridge out of Marin is the'', the elements of the predicted output embedding vector of the model compared to the elements from the Jacobian reconstruction for both the original Jacobian (blue points) and detached Jacobian operations (red points), shown for Qwen 3 14B. Note that the detached Jacobian reconstructions match the predicted embedding, with relative error (the norm of the reconstruction error divided by the norm of the output embedding) less than $10^{-13}$ for double floating-point precision. See reconstructions for Llama 3.2 3B and Gemma 3 4B in Fig. \ref{fig:equivalent_linear_llama_qwen_gemma}.
}\label{fig:arch_and_reconstruction}
\end{figure}

The transformer decoder is the architecture of choice for large language models \citep{vaswani2017attention} and efforts toward a conceptual understanding of its mechanisms are ongoing \citep{sharkey2025open}. Significant insights include sparse autoencoders for conceptual activations in LLMs \citep{bricken2023monosemanticity, templeton2024scaling, lieberum2024gemma}, linear probes \citep{alain2016understanding}, “white-box” alternative architectures \citep{yu2023white} and analytic results on generalization \citep{cowsik2024geometric}. While transformers are  complex globally nonlinear functions of their input, we demonstrate how to compute an equivalent linear system that reconstructs the predicted output embedding for a given input sequence up to double floating-point precision. 

Our approach directly extends the framework of \citet{elhage2021mathematical}, who analyzed attention-only transformers as interpretable linear circuits, but were limited to small models without MLPs (due to gated activation functions) or normalization layers. We show that by detaching nonlinear terms from the gradient computation, modern LLMs with gated activations (as well as softmax attention and normalization) can be decomposed into an equivalent linear system for a given input. Recently, \citet{kadkhodaie2023generalization} showed that powerful image denoising diffusion models with ReLU activations and certain architectural constraints are piecewise linear functions which can be computed via the Jacobian and can be clearly interpreted as low-dimensional adaptive linear filters with comprehensible singular vectors.

For many open-weight LLMs, every component operation (gated activations, attention, and normalization) can be expressed in the form $A(x) \cdot x$, where $A(x)$ represents an input-dependent linear transform and $x$ preserves the linear pathway. The gradient operation with respect to the input can be manipulated at inference by freezing the $A(x)$ terms at their values during inference operation with the detach operation such that the output embedding prediction has a linear equivalent as in Fig. \ref{fig:arch_and_reconstruction}. This ``detached'' Jacobian $\vect{J^+}$ computation captures the complete forward operation of the model, including activation functions and attention modules, although it must be recomputed for each input sequence (as it is not piecewise linear but ``pointwise'' linear).

This approach allows us to analyze a model from input embeddings to predicted output embedding as an equivalent linear system for a particular input sequence. By examining the singular value decomposition (SVD) of the equivalent linear system, we can measure the local dimensionality of the learned manifolds involved in next-token prediction and can decode the singular vectors into output tokens. This analysis can also be done layer by layer, or for individual attention and multilayer perceptron (MLP) modules, in order to observe how these models compose next-token predictions. 

We demonstrate equivalent linearity in model families including Qwen 3, Gemma 3, Llama 3, at a range of sizes up to Qwen 3 14B. (See the appendix for additional equivalent linear demonstrations for Deepseek R1 0528 Qwen 3 8B Distill, Phi 4, Mistral Ministral and OLMo 2). This approach offers a path to interpreting LLMs for specific inputs that could serve as a complement to other powerful interpretability methods. While this is a local method that is somewhat computationally intensive, this approach does not require additional training as required for sparse autoencoders. For example, training sparse autoencoders for Gemma 2 9B (Lieberum et al., 2024) required substantial compute across multiple feature widths and layers, and must be repeated for each new model and layer. Our approach works immediately on LLMs with gated activations and zero-bias linear layers, and produces a more exact representation for interpretation than other methods. If equivalent linear mapping were applied to next-token prediction at scale, this would offer a form of interpretability as the difficult but tractable problem of analyzing many equivalent linear systems.

\section{Method}

\subsection{The Jacobian of a deep ReLU Network}    
\citet{mohan2019robust} observed that deep $ReLU$ networks for image denoising which utilize zero-bias linear layers are  ``adaptive linear'' functions due to their homogeneity of order 1 at a given fixed input, which enables interpretation as an equivalent linear system. Given the homogeneity at a fixed input, the network's output can be reproduced by numerically computing the Jacobian matrix of the network at a particular input image $\vect{x_{im}^{*}}$ and multiplying it by $\vect{x_{im}^{*}}$.  
\begin{equation}\label{eq:deeprelu}
    \vect{y^*_{im}} =  \vect{J}(\vect{x_{im}^{*}})\cdot\vect{x_{im}^{*}} 
\end{equation}
Due to the global nonlinearity of the network, the Jacobian must usually be computed again at every input of interest. The Jacobian may be the same for similar inputs in the same piecewise region of the response \citep{balestriero2021fast, black2022interpreting} (but this will be demonstrated to not be the case for transformer architectures).

\subsection{The Jacobian of a transformer decoder}

Many open weight LLMs also use linear layers with zero bias, as required for linearity in the architecture of \citet{mohan2019robust}. A transformer decoder predicts an output token embedding $\vect{y}$ given a sequence of $k$ input tokens $\vect{t}=(\vect{t_{0}},\vect{t_{1}}...,\vect{t_{k}})$ mapped to input embedding vectors $\vect{x} = (\vect{x_0},\vect{x_1}...,\vect{x_k}$), where $\vect{t^*}$ and $\vect{x^*}$ represent a particular sequence. The output embedding prediction is a nonlinear function of the input embedding vectors $\vect{x_{0}}, \vect{x_{1}}, $...$ \vect{x_{k}}$, as LLMs utilize nonlinear gated activation functions for layer outputs (SwiGLU for Llama 3, GELU for Gemma 3 and Swish for Qwen 3) as well as normalization and softmax attention blocks. 

Gated activations like $Swish(\vect{x}) = \vect{x} \cdot sigmoid(\vect{\beta \cdot x})$, with a linear term and a nonlinear term, are also an ``adaptive'' linear function or, more generally, an adaptive homogeneous function of order 1 \citep{mohan2019robust}. If the $sigmoid(\vect{\beta \cdot x})$ term that gives rise to the nonlinearity is frozen for a specific numerical input, e.g. an embedding vector $\vect{x_0^*}$ \citep{elhage2021mathematical} (or equivalently detached from the computational graph with respect to the input), then we have a linear function valid only at $\vect{x_0^*}$ where (\ref{eq:deeprelu}) holds and we can numerically compute a Jacobian matrix that carries out  $Swish(\vect{x^*_0})$ as a linear operation. 

Below we show that computing the Jacobian after effectively substituting specific values for the nonlinear terms also works for other gated activation functions, normalization layers and softmax attention blocks. We further demonstrate that for a given input sequence we can apply necessary gradient detachments so that the entire transformer decoder is an adaptive homogeneous function of order $1$, and numerically compute the equivalent linear system that reproduces the transformer output embedding $\vect{y^*}$.

The Jacobian $\vect{J(x)}$ of a transformer is the set of matrices generated by taking the partial derivative of the decoder inference function $\vect{y(x)} = f(\vect{x_0},\vect{x_1}...,\vect{x_k})$, with respect to each element of each $\vect{x_i}$ (where $\vect{x_i}$ for Llama 3.2 3B has length $3072$, for example, and therefore the Jacobian matrix for each embedding vector is a square matrix of this size). If a transformer decoder were naturally a homogeneous function of order 1, this Jacobian would generate an equivalent representation of the network.

However, this is not the case. In order to numerically compute an equivalent linear representation, we introduce a ``detached'' Jacobian $\vect{J^+}$, which is a set of matrices that captures the full nonlinear forward computation for a particular input sequence $\vect{x^*}$ as a linear system. The detached Jacobian is the numerical Jacobian of the LLM forward operation when its gradient includes a specific set of $detach()$ operations for the nonlinear terms in the normalization, activation and attention operations that force the function to be ``adaptively'' homogeneous of order 1. The detached Jacobian operates on its corresponding input embedding vector to provide a reconstruction of the LLM forward operation (shown in Fig. \ref{fig:jacobian_sum} and validated in Fig. \ref{fig:arch_and_reconstruction}B by the PyTorch ``allclose'' function for absolute and relative tolerances of $10^{-13}$).
\begin{equation} \label{eq:jacobianpseudo}
\vect{y^*} =  \sum_{i=0}^{k} \vect{J^+_i}(\vect{x^*})\cdot\vect{x^*_{i}} 
\end{equation}

The conventional Jacobian $\vect{J}$ for a particular input sequence $\vect{x^*}$ (as in \citet{mohan2019robust}) does not  generate an accurate reconstruction the nonlinear LLM forward operation since the transformer function is not homogeneous of order $1$. The detached Jacobian $\vect{J^+}$ evaluated at $\vect{x^*}$ is the result of an alternative gradient path through the same network which is homogeneous with respect to the input $\vect{x^*}$. The detached Jacobian $\vect{J^+}$ only generates an accurate reconstruction at $\vect{x^*}$ and not in the local neighborhood due to the strong nonlinearity of the decoder inference function. The detached Jacobian matrices differ for every input sequence and must be computed numerically for every sequence. 

\subsection{Nonlinear layers as linear operators for a given input}
In order to achieve linearity, modifications must be made to the gradient computations of the RMSNorm operation, the activation function (SwiGLU in Llama 3.2) and the softmax term in the attention block output. 

\subsubsection{Normalization}
Normalization layers like LayerNorm \citep{xu2019understanding} or RMSNorm \citep{zhang2019root} with zero bias are nonlinear with respect to their input because they include division by the square root of the variance of the input. 

\begin{equation} \label{eq:rmsnorm0}
norm(\vect{x}) = \frac{\vect{x}}{\sqrt{var(\vect{x})}}
\end{equation}

\citet{mohan2019robust} devised a novel bias-free batch-norm layer which detaches the variance term from the network's computational graph (see their code implementation). Their batch-norm layer returns the same values as the standard batch-norm layer, but it is linear at inference as the nonlinear operation is removed from the gradient computation.  This is also similar to the ``freezing'' of nonlinear terms in attention-only transformers from \citet{elhage2021mathematical}.

We make a  similar change for Llama 3.2 3B by altering how the gradient with respect to the input is computed at inference for RMSNorm. This is accomplished by substituting the value for the input vector $\vect{x^*}$ for only the variance term as in (\ref{eq:rmsnorm1}). In PyTorch, this is accomplished by cloning and detaching the $\vect{x}$ tensor within the variance operation, so its value will be treated as a constant. The gradient operation is still tracked for $\vect{x}$ in the numerator, so that term will be treated as a variable by the PyTorch autograd function for computing the Jacobian. The gradient of the function is then computed at $\vect{x^*}$ (we assume for simplicity an input sequence of length $1$). 

\begin{equation} \label{eq:rmsnorm1}
norm(\vect{x}) = \frac{\vect{x}}{\sqrt{var(\vect{x^*})}}
\end{equation}

We define the detached Jacobian as follows:

\begin{equation} \label{eq:rmsnorm2}
\vect{J^+_{n}} = [\frac{\partial}{\partial \vect{x}}norm(\vect{x})]|_{\vect{x}=\vect{x^*}}
\end{equation}
We can rewrite the pointwise linear RMSNorm as follows:

\begin{equation} \label{eq:rmsnorm3}
norm(\vect{x^*}) = \vect{J^+_{n}}(\vect{x^*})\cdot\vect{x^*}
\end{equation}

At inference for a given input, we now have a linear RMSNorm whose output is numerically identical to the one used in training. However, when we take the gradient with respect to the input vector $\vect{x}$ in $eval$ mode, the numerical output is the detached Jacobian matrix $\vect{J^+_{n}}$, which we can use to reconstruct the normalization output as a linear system.

The goal is to apply this same approach for other nonlinear functions in the decoder such that the entire computation from the input embedding vectors to the predicted output is linear for a given input, and we can compute and interpret the set of detached Jacobian matrices.

\subsubsection{Activation functions}

While \citet{mohan2019robust} relied on $ReLU$ activation functions, which do not require any changes to achieve linearity, Llama 3.2 3B uses SwiGLU \citep{shazeer2020glu}, Gemma 3 uses approximate GELU \citep{hendrycks2016gaussian} and Qwen 3 uses Swish for activation functions. There is a linear $\vect{x}$ term in each of these, and the gradients can be cloned and detached from the nonlinear terms. This manipulation produces a pointwise linear Swish layer with respect to the input $\vect{x}$. 

\begin{equation} \label{eq:swiglu}
 \text{Swish}\left(\vect{x}\right)= \vect{x}\cdot \text{sigmoid}(\beta \cdot \vect{x})
\end{equation}

\begin{equation} \label{eq:swiglu1}
 \text{Swish}\left(\vect{x^*}\right)= \vect{x}\cdot \text{sigmoid}(\vect{\beta \cdot x})|_{\vect{x}=\vect{x^*}} 
\end{equation}

\begin{equation} \label{eq:swiglu2}
 \text{Swish}\left(\vect{x^*}\right)=([\frac{\partial}{\partial \vect{x}}\text{Swish}(\vect{x})]|_{\vect{x}=\vect{x^*}})\cdot\vect{x^*}
\end{equation}

\begin{equation} \label{eq:swiglu3}
 \text{Swish}\left(\vect{x^*}\right)=\vect{J^+_{Swish}}(\vect{x^*})\cdot\vect{x^*}
\end{equation}

Detaching the gradient from the Swish output thus allows for a pointwise linear form of Swish at inference. A similar procedure may be carried out for SwiGLU with Llama 3 and GELU with Gemma 3 (see supplement, eq. \ref{eq:gelu0}).

\subsubsection{Attention}

The softmax operation at the output of the attention block can also be detached, with the linear relationship preserved through the subsequent multiplication with $\vect{V}$, which is a linear function of $\vect{x}$. Below, $\vect{Q} = \vect{W_{Q}}\vect{x}$, $\vect{K} = \vect{W_{\vect{K}}}\vect{x}$ and $\vect{V} = \vect{W_{\vect{V}}}\vect{x}$.

\begin{equation} \label{eq:attn_softmax00}
Attn(\vect{Q}, \vect{K}, \vect{V}) = softmax(\frac{\vect{Q}\vect{K}^T}{\sqrt{d_k}})\cdot\vect{V}
\end{equation}

\begin{equation} \label{eq:attn_softmax0}
Attn(\vect{x}) = [softmax(\frac{\vect{Q}\vect{K}^T}{\sqrt{d_k}})|_{\vect{Q}=\vect{Q^*},\vect{K}=\vect{K^*}}]\cdot\vect{\vect{W_{V}}}\vect{x}
\end{equation}

\begin{equation} \label{eq:attn_softmax1}
Attn(\vect{x^*}) = ([\frac{\partial}{\partial \vect{x}}Attn(x)]|_{\vect{x}=\vect{x^*}})\cdot\vect{x^*}
\end{equation}

\begin{equation} \label{eq:attn_softmax2}
Attn(\vect{x^*}) = \vect{J^+_{Attn}}(\vect{x^*})\cdot\vect{x^*}
\end{equation}

The linear $\vect{x}$ term within $\vect{V}$ makes it possible for the attention block to be pointwise linear at inference, as the gradient for the softmax output is detached.

\subsubsection{The Transformer Decoder}

With the the above gradient detachments for the normalization layers, activation functions and attention blocks, the transformer decoder network is linear with respect to $\vect{x^*}$ when evaluated at $\vect{x^*}$ (shown here for length $k$).

\begin{equation} \label{eq:decoder_jacobian}
\vect{y^*} = \sum_{i=0}^{k}\vect{J}^+_\vect{i}(\vect{x^*})\cdot\vect{x_i^*}
\end{equation}
The output of the network incorporating the above gradient detachments is unchanged from the original architecture but has an equivalent linear representation. 

\section{Results}

\subsection{Pointwise linearity of the predicted output}
In order to validate whether the detached Jacobian achieves reconstruction with a linear representation, we can compare the predicted output embedding vector for a given input token sequence to the reconstruction of the output.  As a baseline, we can also compute the reconstruction using the conventional Jacobian as in \citet{mohan2019robust} and examine its accuracy. Given the above argument that the appropriate gradient detachments are necessary to achieve output reconstruction, we expect the detached Jacobian to accomplish reconstruction, but the conventional Jacobian to fail.

Fig. \ref{fig:arch_and_reconstruction}B compares the network output to both the conventional and detached Jacobian reconstructions for Llama 3.2 3B and Qwen 3 14B. The reconstruction of the output embedding with the detached Jacobian matrices falls on the identity line when compared with the output embedding, showing accurate reconstruction, while the reconstruction with the conventional Jacobian is not at all close to the output. This comparison therefore demonstrates the validity of the reconstruction with the linear system of the detached Jacobian for Qwen 3 14B for a particular input.

In order to examine the fidelity of the detached Jacobian reconstruction, we compared the reconstruction against the network output using PyTorch function allclose with varying tolerance levels. The reconstructions achieved numerical agreement within a relative and absolute tolerance of $10^{-13}$. This tolerance is approximately $50$ times the machine epsilon of $2.2 \cdot 10^{-16}$ for 64-bit floating-point numbers, indicating high-fidelity reconstruction that is numerically equivalent to the reference implementation for practical purposes. As an additional metric, the norm of the detached Jacobian reconstruction error divided by the norm out of the output is on the order of $10^{-14}$.

The numerical computation of the full detached Jacobian matrix takes on the order of 10 seconds for an input sequence of 8 tokens for Llama 3.2 3B in float32 on a GPU with 24 GB VRAM. In contrast, the full Jacobian matrix for the same sequence at float64 precision with Qwen 3 14B on a GPU with 40 GB VRAM takes 20 seconds. An approximate method for computing the top $k$ singular vectors of the detached Jacobian without forming the full matrix utilizing Lanczos iteration has also been implemented in JAX for Gemma 3 4B, allowing for the efficient computation of the top $16$ singular vectors of the detached Jacobian for up to $100$ input-token input. The maximum length tested on a GPU with $80$ GB VRAM was over $400$ tokens for only the top singular vector corresponding to each token. The Lanczos method trades reconstruction precision for scalability while preserving interpretability, and examples are available in the code repository.

\begin{figure*}[!t]
\centering
\includegraphics[width=0.99\textwidth]{"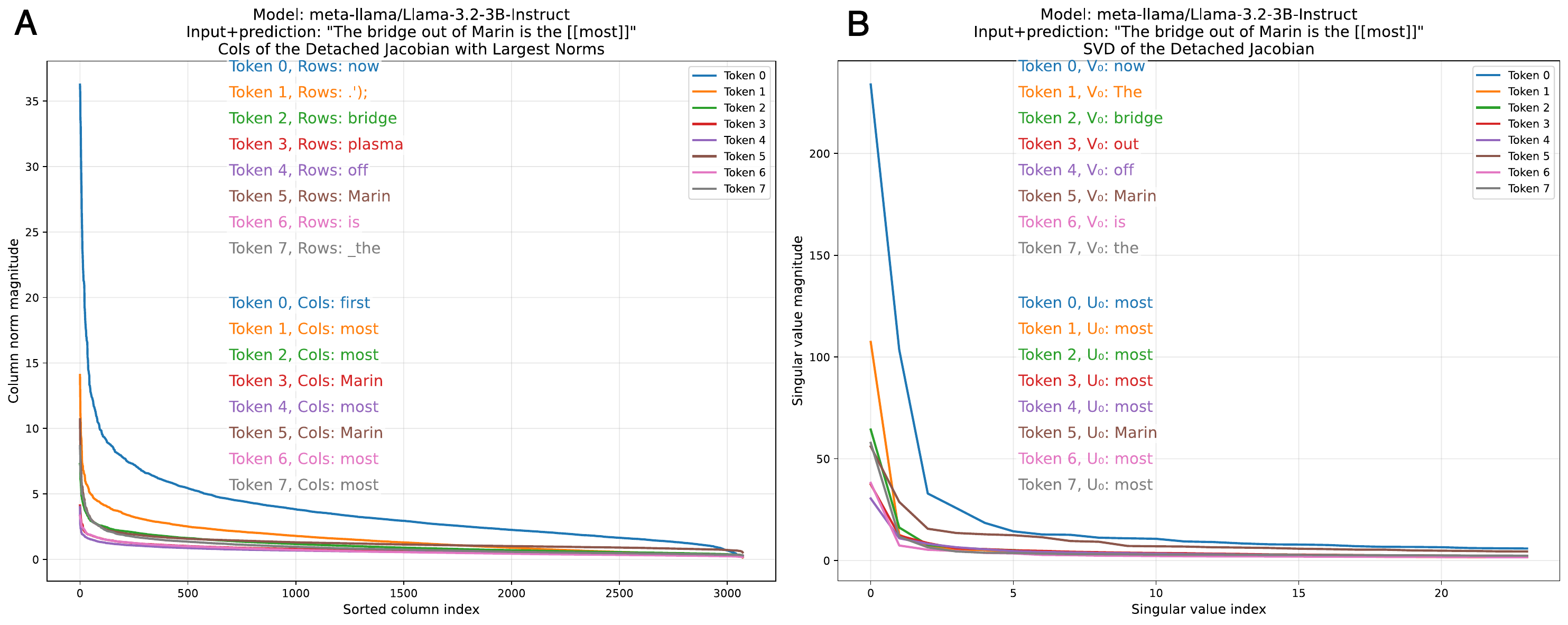"}
\caption{Given the sequence ``The bridge out of Marin is the'', the most likely prediction is ``most'' for Llama 3.2 3B. The detached Jacobian matrices for each token represent an equivalent linear system that computes the predicted output embedding. A) We show the features which drive large responses in single units in the last decoder layer, which are the rows of the detached Jacobian with the largest norm values, and decode each of those into the most likely input embedding token. The block of words at the top shows the ordered decoded ``feature" input tokens from the largest rows of the detached Jacobian matrix for the input tokens. A similar operation is carried out for columns of the largest norm values, which are decoded to the output token space. Note that the activation distribution of column magnitudes is fairly sparse, with only a few units driving the response. B) We take the singular value decomposition of the detached Jacobian matrix corresponding to each input token, which summarizes the modes driving the response, and decode the right and left singular vectors $V$ and $U$ to input and output embeddings, shown in colors. The singular value spectrum is extremely low rank, and decoding the $U$ singular vectors returns candidate output token, including ``most'' and ``first''. Decoding the $V$ singular vectors returns variants of the input tokens like ``bridge'', ``Marin'' and ``is'', as well as others that are not clearly related to the input sequence.
}\label{fig:singular_vectors}
\end{figure*}

\subsection{Single-unit feature selectivity and invariance}
Since the detached Jacobian applied to the input embedding reproduces the predicted output embedding vector, and the elements of the predicted output embedding vector are the units of the last transformer layer, the rows of the detached Jacobian matrices represent the input features to which the last layer units are selective and invariant for that particular input sequence \citep{kadkhodaie2023generalization,mohan2019robust}.

The activation of a particular unit in the last layer is determined by the inner product of a row of the detached Jacobian and the input embedding vector. We can sort by the magnitude of row norms, then map the largest-magnitude rows of the detached Jacobian back to the input embedding space (via cosine similarity to the input embedding matrix, since input embeddings are not typically mapped back to tokens during normal model operation) to determine the tokens that cause each unit to be strongly positive or negative. We can see in Fig. \ref{fig:singular_vectors}A that the units respond strongly to the words of the prompt, including ``bridge'', ``Marin'' and ``is''. Decoding of the rows of the detached Jacobian for each token as well as the distribution of activations for this sequence is shown in Fig. \ref{fig:singular_vectors}A. The columns of the Jacobian can also be decoded in the conventional manner to the output token space with the unembedding layer, and these turn out to be tokens that could be predicted, which include words like ``most'' or ``first'', which could be acceptable outputs.

\subsection{Singular vectors of the detached Jacobian}

An alternative approach is to look at the singular value decomposition of the detached Jacobian $\vect{J^+_i} = \vect{U}\vect{\Sigma}\vect{V^T}$, following \citet{mohan2019robust}. Since the detached Jacobian represents the forward computation, the fact that the SVD is very low rank shows the entire forward computation can be approximated with only a few singular vectors operating on the input embeddings.

Unlike image denoising models \citep{mohan2019robust, kadkhodaie2023generalization} where input and output spaces are similar and singular vectors U and V have a high cosine similarity, corresponding left and right singular vectors of LLMs differ substantially. This reflects the asymmetric nature of next-token prediction, as right singular vectors V capture which input token features drive the computation, while left singular vectors U capture which output token directions are predicted.

In Fig. \ref{fig:singular_vectors}B, the singular vectors are decoded for Llama 3.2 3B (and for other models in supplemental Fig. \ref{fig:singular_vectors_across_models}). The right singular vectors $V$ are decoded to input tokens in the same way the rows of the detached Jacobian were above (nearest-neighbor to input embeddings from cosine similarity), and we see similar decoding of the top tokens to the features driving the most active single units. The left singular vectors $U$ can be decoded to output embedding tokens (with the conventional method from the unembedding matrix), and ``most'' is the strongest, as it was in the columns of the detached Jacobian matrices.

\definecolor{highlightnone}{RGB}{255,255,255}  
\definecolor{highlight0}{RGB}{255,255,0}      
\definecolor{highlight1}{RGB}{255,160,160}  
\definecolor{highlight2}{RGB}{0,255,0}      
\definecolor{highlight3}{RGB}{255,165,0}    
\definecolor{highlight4}{RGB}{173,216,230}  
\definecolor{highlight5}{RGB}{221,160,221}  

\newcommand{\hlnone}[1]{\colorbox{highlightnone}{#1}}
\newcommand{\hlzero}[1]{\colorbox{highlight0}{#1}}
\newcommand{\hlone}[1]{\colorbox{highlight1}{#1}}
\newcommand{\hltwo}[1]{\colorbox{highlight2}{#1}}
\newcommand{\hlthree}[1]{\colorbox{highlight3}{#1}}
\newcommand{\hlfour}[1]{\colorbox{highlight4}{#1}}
\newcommand{\hlfive}[1]{\colorbox{highlight5}{#1}}
\newcommand\ChangeRT[1]{\noalign{\hrule height #1}}

\begin{table*}[!ht]
\begin{adjustbox}{width=\textwidth,center}
    \centering \scriptsize
    \begin{tabular}{|l|l|l|l|}
    \hline
 ~ & Input token 0 & Input token 1 & Input token 2 \\ \hline
  \ChangeRT{1.4pt}
    Layer 25\_0 & largest  \hlfive{most}  first  longest  latest  fastest  last  third & \hlzero{bridge}  \hlzero{bridges}     \hlzero{Bridge}  gateway & hardest  ones  \hlthree{exit}  easiest  first  \hlfive{most}  fastest  \hltwo{highway} \\ \hline
 Layer 25\_1 & \hlzero{bridge}     \hlzero{bridges}    \hlzero{Bridge}  \hlzero{Bridges}  brid & \hlzero{bridges}  \hlzero{bridge}  \hlzero{Bridge} \hlzero{bridge}  \hlzero{Bridge} & (\hlthree{exit} \hlthree{exit}  exits  eternity .\hlthree{exit} \\ \hline
 Layer 25\_2 & \hlzero{bridges}  \hlzero{bridge}  \hlzero{Bridge}  \hlzero{bridge}   parliament & \hlthree{Exit}  \hlthree{exit}  jams & INCIDENT      symbolism \\ \hline
 \ChangeRT{1.4pt}
    Layer 26\_0 & first  \hlfive{most}  largest  last  longest  latest  gateway  \hlone{only} & \hlzero{bridge}  \hlzero{bridges}    metaphor  gateway  connecting & \hltwo{highway}  first  \hlthree{exit}  ones  last   hardest  roads \\ \hline
 Layer 26\_1 & \hlzero{bridge}  \hlzero{bridges}    metaphor  \hlzero{Bridges}  \hlzero{Bridge} & \hlzero{bridges}   \hlzero{bridge}  structures   brid \hlzero{bridge} & .charset    jams Margins \\ \hline
 Layer 26\_2 & parliament  structures   \hlzero{bridges}  Parliament    \hlzero{bridge} & \hlthree{Exit}   \hlthree{exit}  choke  \hlthree{Exit}  panicked & symbolism    metaphor \\ \hline
 \ChangeRT{1.4pt}
    Layer 27\_0 & first  last  largest  \hlzero{bridge}  longest  \hlfive{most}  oldest  latest & \hlzero{bridge}  \hlzero{bridges}     \hlzero{Bridge}   \hlzero{Bridges} & last  first  \hlthree{exit}  \hltwo{highway}  bottleneck  next  road  choke \\ \hline
 Layer 27\_1 & \hlzero{bridge}   \hlzero{bridges}     \hlzero{Bridge}  \hlzero{Bridges} & \hlzero{bridges}   \hlzero{bridge}  \hlzero{Bridge} \hlzero{bridge}  brid \hlzero{Bridge} & EXIT \hlthree{exit}  exits   (\hlthree{exit} \\ \hline
 Layer 27\_2 & \hlzero{bridge} \hlzero{bridge}  \hlzero{bridges}   \hlzero{Bridge}    structures & \hlthree{Exit} \hlthree{exit}  \hlthree{Exit} & \hlthree{exit} .\hlthree{exit} incident  EXTRA incidents \\ \hline
 \ChangeRT{1.4pt}
    Layer 28\_0 & \hlzero{bridge}  longest  largest  first  busiest  last  oldest  \hlfive{most} & \hlzero{bridge}  \hlzero{bridges}     \hlzero{Bridge}  \hlzero{Bridge} & \hltwo{highway}  \hlthree{exit}  bottleneck  highways  \hltwo{Highway}  last  road  exits \\ \hline
 Layer 28\_1 & \hlzero{bridge}  \hlzero{bridges}     \hlzero{Bridge}  \hlzero{Bridge} & \hltwo{highway}  highways   coast  freeway  roads  road  route & \hlthree{exit}  exits    EXIT   \hlthree{exit} \hlthree{Exit} \\ \hline
 Layer 28\_2 & \hlzero{bridge} \hlzero{bridge}  \hlzero{bridges}    \hlzero{Bridge}  brid & \hlthree{Exit}  \hlthree{exit}  \hlthree{Exit}  \hlthree{exit}  \hlthree{exit} & \hlthree{exit} Saddam  Mosul  Kuwait incident  metaphor \\ \hline
 \ChangeRT{1.4pt}
    Layer 29\_0 & \hlzero{bridge}  \hlone{only}  fourth  last  third  longest  fifth  \hlfive{most} & \hlzero{bridge}  \hlzero{bridges}    \hlzero{Bridge}    \hlzero{Bridges} & \hlone{only}  last  first  \hltwo{highway}  third  highways  \hlthree{exit}  fourth \\ \hline
 Layer 29\_1 & \hlzero{bridge}  \hlzero{bridges}     \hlzero{Bridge}  \hlzero{Bridges} & coast  \hltwo{highway}  road  driveway  coastline  roads  highways  freeway & exits \hlthree{exit}    EXIT \\ \hline
 Layer 29\_2 & \hlzero{bridge}  \hlzero{bridges} \hlzero{bridge}   structures  brid   structure & \hlthree{Exit} \hlthree{exit}   \hltwo{Highway}  \hlthree{Exit} & Saddam    Mosul  Elvis  metaphor  incident \\ \hline
 \ChangeRT{1.4pt}
    Layer 30\_0 & \hlzero{bridge}  \hlfive{most}  longest  fourth  third  last  \hlone{only}  fifth & \hlzero{bridge}  \hlzero{bridges}    \hlzero{Bridge}   \hlzero{Bridge} & \hltwo{highway}  \hlone{only}  \hlzero{bridge}  last  first  \hltwo{Highway}  road  highways \\ \hline
 Layer 30\_1 & \hlzero{bridge}   \hlzero{bridges}     \hlzero{Bridges}   \hlzero{Bridge} & coast  freeway   \hltwo{highway}  coastline  road  roads  highways & \hlzero{bridge}   \hlzero{Bridge}  \hlzero{bridges} \hlzero{bridge}   brid \\ \hline
 Layer 30\_2 & \hlzero{bridge}  structure  structures  \hlzero{bridges} \hlzero{bridge}   brid & sail  seab    sailing  Bermuda  ship & Memphis  Kuwait  Jordan   Saddam   Iowa \\ \hline
 \ChangeRT{1.4pt}
    Layer 31\_0 & \hlzero{bridge}  \hlfive{most}  \hlone{only}  last  longest  first  third  largest & \hlzero{bridge}   \hlzero{bridges}  \hlzero{Bridge}   \hlzero{Bridge} & \hlone{only}  last  \hltwo{highway}  first  \hlzero{bridge}  \hlthree{exit}  \hltwo{Highway}  \hlfive{most} \\ \hline
 Layer 31\_1 & coast  airlines Interior  airline  interior  Lua  Speedway & coast  coastline  coastal   Coast   Coastal  route & \hlzero{bridge}   \hlzero{Bridge}  \hlzero{bridges}  \hlzero{bridge}  underwater  brid \\ \hline
 Layer 31\_2 & \hlzero{bridge}   \hlzero{bridges} \hlzero{bridge}  brid  \hlzero{Bridge}   structure & ship    sail   sailing  dock  seab & Jordan  Memphis  Kuwait      Mississippi \\ \hline
 \ChangeRT{1.4pt}
    Layer 32\_0 & \hlzero{bridge}  \hlfive{most}  \hlone{only}  first  last  longest  third  largest & \hlzero{bridge}   \hlzero{Bridge}  \hlzero{bridges}  \hlzero{Bridge}  \hlzero{bridge} & \hlone{only}  last  first  \hltwo{highway}  \hlfive{most}  main  route  \hlthree{exit} \\ \hline
 Layer 32\_1 & interior  airline  steam  airlines Trail  breed   vacuum & coast  coastal  coastline  route   Coast  Route  beach & \hlzero{bridge}  span  underwater  connecting  deck  public   member \\ \hline
 Layer 32\_2 & \hlzero{bridge} \hlzero{bridge}   \hlzero{bridges}   \hlzero{Bridge}  brid \hlzero{Bridge} & ship  sail    dock  sailing   seab & Kuwait    Jordan  Memphis  Edmonton  Mississippi  Nile \\ \hline
 \ChangeRT{1.4pt}
    Layer 33\_0 & \hlone{only}  first  last  \hlfive{most}  third  main  second  subject & \hlzero{bridge}  \hlzero{Bridge}   \hlzero{bridges} \hlzero{Bridge}   \hlone{only} & \hlone{only}  last  first  key  main  same  \hlfive{most}  \hlthree{exit} \\ \hline
 Layer 33\_1 & planet   interior  cabin  floors  roots & coast  coastline  coastal   Coast  route  beach  Coastal & span  public  member  library  platform  floating  intervening  deck \\ \hline
 Layer 33\_2 & \hlzero{bridge} \hlzero{bridge}   structure  \hlzero{bridges}  brid  \hlzero{Bridge} & ship   orbit   aircraft  sail  vessel & Kuwait    Nile  Edmonton  Saskatchewan  Tulsa \\ \hline
    \end{tabular}
\end{adjustbox}
    \caption{The top eight tokens decoded from the largest three singular vectors of the detached Jacobian for the layer outputs from Qwen 3 14B for the sequence ``The bridge out of Marin is the'' with the prediction [[only]].  Legend: \hlzero{``Bridge''}, \hlone{``only''}, \hltwo{``highway''}, \hlthree{``exit''}, \hlfive{``most''}. Semantic concepts emerge clearly by layer 25. The predicted token 'only' appears prominently in later layers alongside related infrastructure and geographic concepts. Note the progression from general bridge concepts in early layers to specific architectural terms (span, deck, platform, floating), geographic terms (coast, coastline, route, beach) and locations with notable bridges in the final layer. See also supplemental Tables \ref{tab:llama-layers-full}, \ref{tab:gemma-layers-full} and \ref{tab:qwen-layers-full} for the longer tables for Llama, Gemma and Qwen.}
\label{tab:qwen3_jacobian_decoded_3cols}
\end{table*}

\subsection{Comparative Analysis of Singular Vectors in Llama 3 and Qwen 3}

A direct comparative analysis of the singular vectors derived from the detached Jacobian matrices of Llama 3 3.2B and Qwen 3 4B offers a lens through which to view not only the shared computational principles of modern LLMs but also their distinct data-driven approaches. While both models demonstrate a consistent hierarchical organization of their predictive computations, they diverge significantly in their semantic richness, their approach to multi-lingual representations, and their tokenization strategies. These differences are made visible by the SVD of their equivalent linear mappings and reveal unique styles that likely reflect their underlying training datasets.

\begin{figure*}[!t]
\centering
\includegraphics[width=0.99\textwidth]{"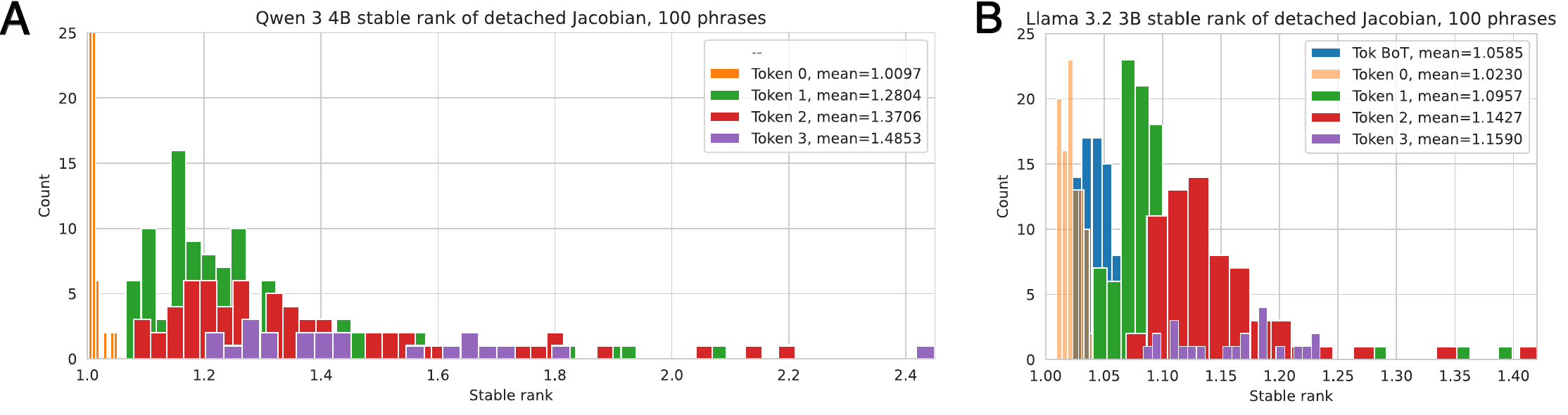"}
\caption{For 100 short input phrases, the stable rank distribution as a function of input token number. Note that Llama 3.2 3B uses a $<|BoT|>$ token and Qwen 3 4B does not.
}\label{fig:rank-100ex}
\end{figure*}

In terms of their singular value spectra over 100 examples, Fig. \ref{fig:rank-100ex} shows that both Llama 3 and Qwen 3 are consistently low-rank. The first token for Qwen 3 has a low average rank at $1.01$ than Llama 3 at $1.06$, but Qwen's next singular vectors are all higher rank than those of Llama. Llama's ``beginning of text'' token is surprisingly of lower rank than the first text token.

In terms of the semantic content of the singular vectors, both Llama 3 and Qwen 3 employ a similar hierarchical strategy. The first singular vector $U_0$ with largest magnitude establishes the foundational layer of prediction. This vector primarily contains high-frequency tokens that provide grammatical structure or represent the most probable continuations. For example, in ``Should have known,'' both models place ``better'' and common punctuation in their $U_0$ vectors. This shared pattern reinforces the hypothesis that the dominant computational axis in transformers is dedicated to establishing a coherent structural and high-probability scaffold upon which more nuanced semantic meaning can be built. See section \ref{examples-comparative} in appendix for more examples of each of these analyses. \textbf{21 phrases out of 100} fit this category.

\subsection*{Llama 3 (Abstract Semantics) vs. Qwen 3 (Direct Semantics)}

A distinction in semantic processing is pronounced in the secondary singular vectors ($U_1$ and $U_2$). Llama 3 consistently demonstrates a rich and abstract English-centric semantic space. For the input ``Will break,'' its $U_1$ vector contains a diverse set of conceptual possibilities like ``confidentiality,'' ``independence,'' ``promises,'' and ``ground.'' This indicates a capacity to reason about abstract concepts that can be ``broken.'' Qwen 3's vectors for the same phrase are more direct and action-oriented, featuring tokens like ``ties,'' ``neck,'' and ``dance,'' alongside Chinese characters for ``stiff'' and ``can't.'' This highlights Llama 3's deep modeling of the nuances and abstractions within the English language.  \textbf{14 phrases out of 100} fit this category.

\subsection*{Qwen 3's Multilingual Reasoning}

Perhaps the most obvious difference revealed by this analysis is Qwen 3's  multilingual and cross-lingual representation capability, which is largely absent in Llama 3's vectors for the analyzed English prompts. In nearly every example, Qwen 3's secondary vectors are populated with non-English tokens—primarily Chinese, but also Russian and others that are conceptually related to the input phrase. For ``The broken,'' Qwen 3's $U_1$ vector includes Chinese tokens for ``bicycle,'' ``vase,'' ``necklace,'' and ``window''—all concrete examples of breakable objects. This demonstrates that Qwen 3 does not operate in a constrained linguistic space; rather, it accesses a unified, cross-lingual conceptual representation to generate predictions. \textbf{38 phrases out of 100} fit this category.

\subsection*{Examples of sub-word Fragments in Qwen 3}

We also observed a difference in tokenization and morphological strategy. Qwen 3's secondary vectors frequently contain what appear to be sub-word fragments or tokenization artifacts (e.g., ``e,'' ``eer,'' ``ection,'' ``ing''). The persistent recurrence of these tokens, often in the $U_2$ vector, suggests that part of Qwen 3's computational process involves constructing or modifying words at a morphological level. This could be an efficient mechanism for handling its  multilingual vocabulary. Llama 3 tends to operate with whole-word semantic tokens, indicating a different approach to vocabulary representation. \textbf{33 phrases out of 100} fit this category.

\subsection{Layer output singular vectors}

Table \ref{tab:qwen3_jacobian_decoded_3cols} shows the top eight tokens decoded from the largest three singular vectors of the detached Jacobians of selected layer outputs for Qwen 3 14B. The words ``bridge'' (and its variants), ``highway'', ``exit'', ``most'' and ``only'' are highlighted to show their appearances in decoded singular vectors. Early layers are excluded as the tokens are unintelligible. The emergence of intelligible tokens in later layers is shown in the tables as something like a phase change in the representation. Qwen 3 generates infrastructure and engineering related concepts before producing ``only''.

\begin{figure*}[!t]
\centering
\includegraphics[width=0.99\textwidth]{"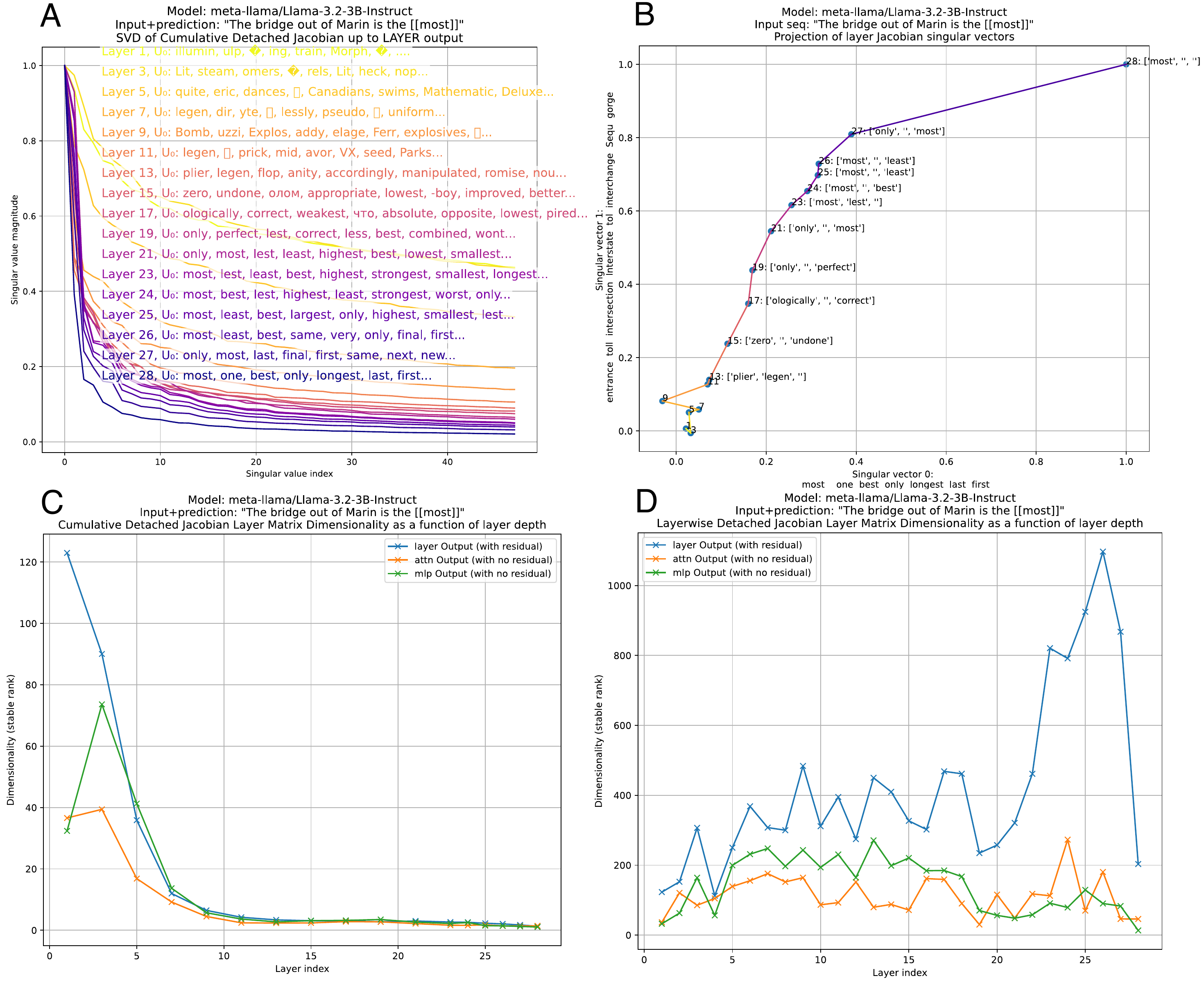"}
\caption{Since the transform representing the model forward operation is linear after detachment, we can also decompose each transformer layer as a linear operation as well. A) The singular value spectrum for the cumulative transform up to layer $i$.  Note that later layers are lower rank than earlier layers. The top singular vectors of the later layers show a clear relation to the prediction of ``most''. B) The projection of the top two singular vectors onto the top two singular vectors of the final layer. The singular vectors of the first 10 layers are very different than those of the last layer, so the projections remain close to the origin. At layer 11, they begin to approach those of the output layer. C) A measurement of the dimensionality of the cumulative transform up to the output of each layer as the stable rank. Within each layer, the outputs of the attention and MLP modules (prior to adding the residual terms) can also be decomposed as linear mappings. The dimensionality decreases deeper into the network at each of these points, except for a slight increase for the attention and MLP module outputs in layer 3. D) The dimensionality of the detached Jacobian for the layer-wise transform at layer $i$ for the layer output, as well as the attention module output and MLP module output. 
}\label{fig:singular_vectors_layers}
\end{figure*}

Fig. \ref{fig:singular_vectors_layers}A shows the normalized singular value spectra of the detached Jacobian at the output of every layer. Llama 3.2 3B has $28$ layers, and decoding the largest singular vectors shows that the word representation of these intermediate operations is not  interpretable until later layers. From the decoding of the top singular vector by layer, ``only'' emerges in layer 19. From the map of the progression of the projection of the top two singular vectors onto the top two singular vectors of the last layer in Fig. \ref{fig:singular_vectors_layers}B, we first see a shift at layer 11 toward the prediction.

Since the layer-by-layer operations are only linear, the stable rank $R = (\sum_{i}^L{S^2_i}) / {S^2_{max}}$ serves as a measure of the effectively dimensionality of the subspace of the representation at a particular layer.

When looking at $W_{0\_\text{to}\_k}$, the cumulative layer transform up through layer $k$, the dimensionality of the detached Jacobian steadily decreases. When considering each layer $i$ as its own individual transform $W_{i}$ (where $ W_{0\_\text{to}\_k} = \prod_{i=0}^k{W_i}$ for the simplified scenario of a single input token; there are other cross-token terms not shown here for mid-layer detached Jacobians for longer input sequences), we also see a large peak in dimensionality near the end. 

\begin{table*}[!ht]
{\centering \small
\begin{tabular*}{0.9775\textwidth}{|p{0.17\textwidth}|p{0.17\textwidth}|p{0.17\textwidth}|p{0.17\textwidth}|p{0.17\textwidth}|}

\hline
Model & Layer intervention & Input sequence & Normal response & Steered response\\
\hline
Llama 3.1 8B & 24 / 36 & `I'm going to arizona to see the' & `I'm going to arizona to see the Grand Canyon. I've heard it's a must see. I've also heard it's a bit of a trek to'  & `I'm going to arizona to see the Grand Canyon, and I'm planning to hike the Bright Golden Gate Bridge (I think that's the name of the trail) in the Grand Canyon.'  \\
\hline
Qwen 3 8B & 24 / 36 & `Here is a painting of the' & `Here is a painting of the same scene as in the previous question, but now the two people are standing on the same side of the building. ' &
 `Here is a painting of the Golden Gate Bridge in San Francisco. The Golden Gate Bridge is one of the most famous bridges in the world. ' \\
\hline
Gemma 3 12B & 33 / 48 & `I went to new york to see the' & `I went to new york to see the memorial and museum. It was a very moving and emotional experience.' & `I went to new york to see the 10th anniversary of the Broadway show, ``The Golden Gate Bridge Bridge.'' It was a great show.' \\
\hline
\end{tabular*}
\caption{Detached Jacobian matrices as steering operators, pilot results with Llama 3.1 8B, Qwen 3 8B and Gemma 3 12B.}
\label{tab:steering}
}\par
\end{table*}

\subsection{The detached Jacobian as a conceptual steering operator}

Steering vectors are a well-known technique for altering LLM outputs \citep{liu2023context} where a vector with certain properties is added to a mid-layer representation, and the sum is passed through the rest of the network to generate an output token. Here we utilize the detached Jacobian as an operator instead of an additive vector, and compute it from an intermediate layer for a ``steering'' phrase like ``The Golden Gate'' (after the ``Golden Gate Claude'' demo \citep{templeton2024scaling}). The model predicts ``Bridge'', and this detached Jacobian matrix is used to steer the continuation of a new phrase toward this concept. For a new input phrase, like ``Here is a painting of the'', the ``new'' input sequence's embedding vectors $\vect{x_{new}^*}$ are multiplied by the detached Jacobian previously computed from the steering concept $\vect{J_L^+}(\vect{x^*_{steer}})$, scaled by $\lambda$ and added to the layer activation $\vect{f_{Li}}$ from the ``new'' input. 

\begin{equation} \label{eq:steering}
\vect{f_{Li}}(\vect{x}) = \lambda\cdot \vect{f_{Li}}(\vect{x_{new}^*}) + (1-\lambda)\cdot \vect{J_{Li}^+}(\vect{x^*_{steer}})\cdot\vect{x^*_{new}}
\end{equation}

This steered intermediate representation is then put through the remaining layers of the network and the next token is decoded. The detached Jacobian must only be computed once for the steering concept, and therefore this method is rather efficient. Table \ref{tab:steering} shows how the detached Jacobian from an intermediate layer imposes the Golden Gate Bridge as the semantic output coherent with the rest of the input sentence, even when it is difficult to make a logical connection. Beyond demonstrating practical utility, the success of the steering operator provides validation that the detached Jacobian captures actual semantic representations. 

\section{Discussion}

The detached Jacobian approach allows for linear representations of the transformer decoder to be found for each input sequence, without changing the output. The intermediate outputs of each layer and the attention and MLP modules are also  accurately reproduced by the detached Jacobian function. 

The detached Jacobian operation is accurate only at the specific operating point at which the matrices were computed by the PyTorch autograd function. A short distance away in the input embedding neighborhood, the detached Jacobian will be extremely different because the manifold is highly curved. (Although local neighborhood validity is less applicable to LLMs which map tokens to embedding vectors, as inputs will only ever discretely sample the embedding space, and there is not an obvious need for exploring the local neighborhood to embedding vectors that do not represent words from the input vocabulary). The manifold is not piecewise linear, but only has a linear equivalent at the operating point, which can be found numerically for every input sequence.

\section{Conclusion}
While our current analysis covers a limited range of examples, the approach suggests a path toward large-scale interpretability by computing the detached Jacobian for many token predictions in a given dataset and analyzing the resulting linear systems to understand semantic patterns across diverse contexts. Given the low-rank nature of the detached Jacobian, our Lanczos method, which efficiently computes only the top singular vectors of the Jacobian, is a step toward making this practical. Future work should explore this scaling potential, moving toward comprehensive equivalent linear analysis of LLM behavior across tasks, domains, and model architectures.

\clearpage

\bibliography{main}
\bibliographystyle{tmlr}

\newpage
\appendix
\onecolumn

\section{Appendix}
\label{sec:appendix}

\renewcommand{\thefigure}{A\arabic{figure}}
\renewcommand{\theHfigure}{A\arabic{figure}}
\setcounter{figure}{0}

\renewcommand{\thefigure}{A\arabic{figure}}

\setcounter{figure}{0}

\subsection{Code availability}

Code is provided as a zip file (and will be made available on github).

\subsection{Pointwise linear GELU}
Gemma 3 uses the approximate GELU activation function. Below $\gamma = 0.44715$. Here is the derivation of the pointwise linear version of GELU used for Gemma 3 in the preceding analysis.

\begin{equation} \label{eq:gelu0}
 \text{GELU}\left(\vect{x}\right)=\frac{1}{2}\vect{x}\left(1+\tanh\left[\sqrt{2/\pi}\left(x + \gamma \vect{x}^{3}\right)\right]\right)
\end{equation}

\begin{equation} \label{eq:gelu1}
 \text{GELU}\left(\vect{x}\right)=\frac{1}{2}\vect{x}\left(1+\tanh\left[\sqrt{2/\pi}\left(x + \gamma \vect{x}^{3}\right)\right]\right)|_{\vect{x}=\vect{x^*}}
\end{equation}

\begin{equation} \label{eq:gelu2} 
 \text{GELU}\left(\vect{x^*}\right)=([\frac{\partial}{\partial \vect{x}}\text{GELU}(x)]|_{\vect{x}=\vect{x^*}})\cdot\vect{x^*}
\end{equation}

\subsection{Singular vectors across model families}

\begin{figure*}[!t]
\centering
\includegraphics[width=0.99\textwidth]{"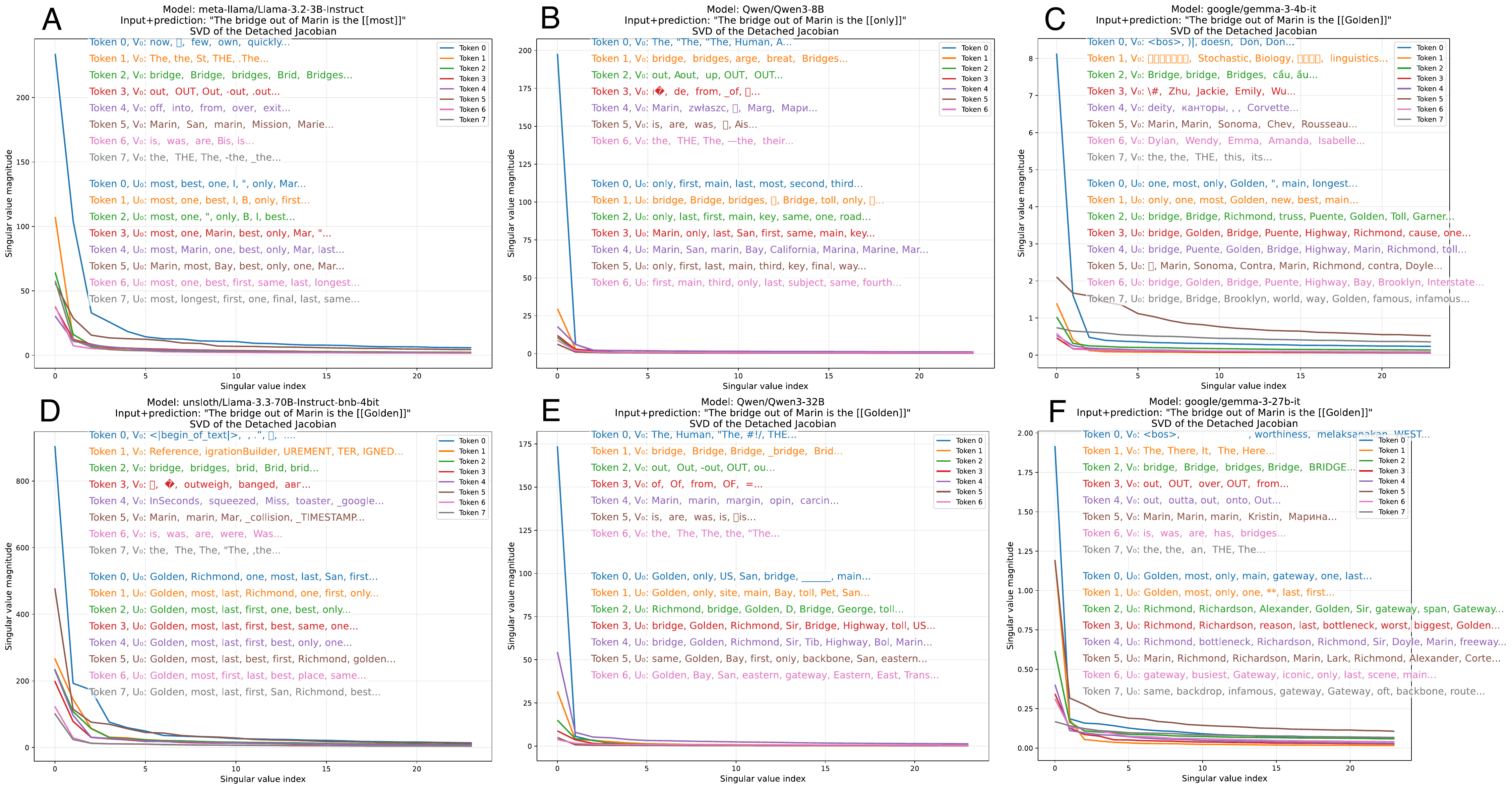"}
\caption{Singular value decomposition of the detached Jacobian for different families and sizes of language models (from $3$B to $70$B parameters) evaluating the input sequence ``The bridge out of Marin is the'', followed by a predicted token. The left singular vectors decode to tokens related to bridges and local geography, particularly the Golden Gate Bridge, while singular value spectra all have extremely low rank (see below for quantification). Each row shows top tokens associated with different singular vectors, demonstrating how models encode semantic knowledge about the input sequence and the prediction. See Fig. \ref{fig:deepseek-8b-distill} for Deepseek R1 0528 Qwen 3 8B Distill, Phi 4, Mistral Ministral and OLMo 2.}\label{fig:singular_vectors_across_models}
\end{figure*}

Fig \ref{fig:singular_vectors_across_models} shows this same analysis for Llama 3, Qwen 3 and Gemma 3 across two different sizes of each. Note the low-rank structure of each of the detached Jacobians, as well as the differing decoding of the top singular vectors from each input embedding vector. The first or ``beginning of sequence'' token has the highest magnitude in each spectrum reflecting how the positional encoding is entangled with semantic information in the detach Jacobian representation.

\subsection{Additional models}
\begin{figure}[!t]
\centering
\includegraphics[width=0.8\columnwidth]{"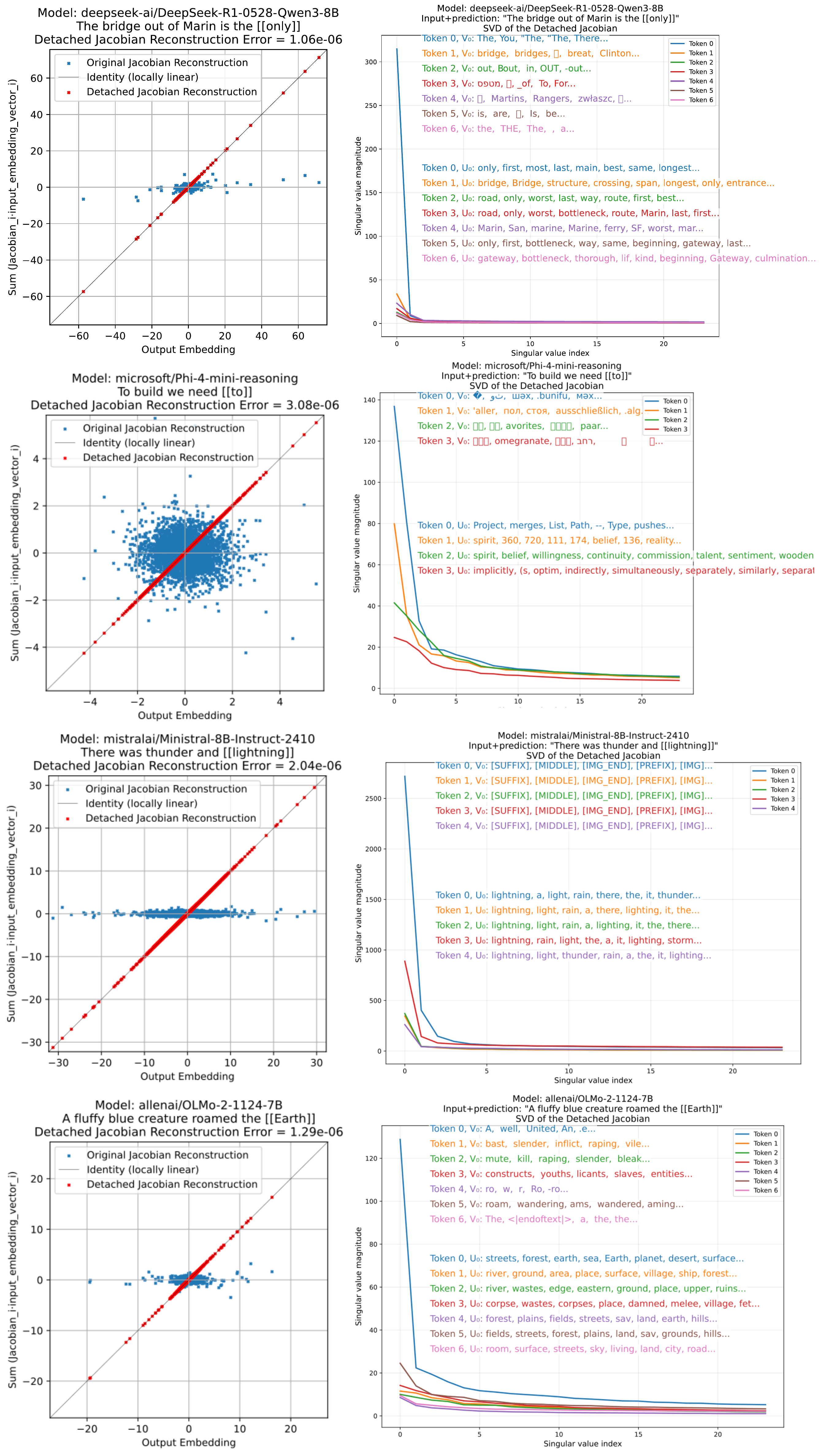"}
\caption{The detached Jacobian reconstruction error and SVD for Deepseek R1 0528 Qwen 3 8B, Phi 4 Mini 4B, Mistral Ministral 8B and OLMo2 7B.
}\label{fig:deepseek-8b-distill}
\end{figure}

Pointwise linearity for Deepseek R1 0528 Qwen 3 8B Distill, Phi 4, Mistral Ministral and OLMo 2 are shown on the following page. See Fig. \ref{fig:deepseek-8b-distill}.


\subsection{Examples for comparative analysis of singular vectors in Llama 3 and Qwen 3}
\label{examples-comparative}



\subsubsection*{Shared High-Probability Tokens in $U_0$}

This pattern shows both models using their primary singular vector ($U_0$) to establish a foundation of common, structurally likely next words.

For the phrase ``To see,'' both models prioritize articles and question words.
\begin{itemize}
    \item \textbf{Qwen 3 $U_0$}: \texttt{the a this an all how what and}
    \item \textbf{Llama 3 $U_0$}: \texttt{the a , and what an if}
\end{itemize}

For ``To complete,'' both models identify determiners as the most probable continuations.
\begin{itemize}
    \item \textbf{Qwen 3 $U_0$}: \texttt{the a this his an my your}
    \item \textbf{Llama 3 $U_0$}: \texttt{this , the a and an (}
\end{itemize}

For ``The final result,'' the $U_0$ vectors in both models are dominated by common prepositions and linking verbs that would grammatically follow the phrase.
\begin{itemize}
    \item \textbf{Qwen 3 $U_0$}: \texttt{of is in for from after , was}
    \item \textbf{Llama 3 $U_0$}: \texttt{, of ... ( is in and}
\end{itemize}

Both models use their primary singular vector ($U_0$) to propose very similar sets of common, structurally-likely next words. This highlights a shared foundational strategy of prioritizing grammatical coherence.

\textbf{21 phrases out of 100} fit this category.

\begin{itemize}
    \item \textbf{Before they:} Both suggest verbs like \texttt{were, can, could, start}.
    \item \textbf{While walking:} Both suggest prepositions of movement like \texttt{in, through, on, along, around}.
    \item \textbf{To see:} Both prioritize articles (\texttt{the, a}) and question words (\texttt{what, how}).
    \item \textbf{Will break:} Both suggest particles like \texttt{down} and \texttt{up}, and articles like \texttt{the, a}.
    \item \textbf{Must leave:} Both list determiners (\texttt{the, a, this}) and prepositions (\texttt{in, at}).
    \item \textbf{Should take:} Both include \texttt{a, the, into,} and \texttt{care}.
    \item \textbf{After reading:} Both list \texttt{the, this, a, about,} and \texttt{"}.
    \item \textbf{When finished:} Both suggest \texttt{,} and \texttt{with}.
    \item \textbf{To begin:} Both prioritize \texttt{,} and \texttt{with}.
    \item \textbf{May open:} Both suggest \texttt{a, the, up,} and \texttt{in}.
    \item \textbf{Could drive:} Both include \texttt{a, the, in,} and \texttt{,}.
    \item \textbf{During lunch:} Both list \texttt{,, time,} and \texttt{break}.
    \item \textbf{To learn:} Both prioritize \texttt{the, more, about,} and \texttt{how}.
    \item \textbf{The green:} Both include \texttt{and, ,, light, is}.
    \item \textbf{The old man:} Both list linking verbs (\texttt{was, is}) and conjunctions (\texttt{and}).
    \item \textbf{To build they:} Both suggest modal verbs (\texttt{have, need, must, would}).
    \item \textbf{The fast car:} Both include \texttt{is, has, and, ,}.
    \item \textbf{The tall building:} Both list \texttt{is, in, has, with}.
    \item \textbf{To create:} Both prioritize articles \texttt{a, an, the}.
    \item \textbf{The response:} Both include \texttt{to, is, of}.
    \item \textbf{The solution:} Both list \texttt{to, of, is, for}.
\end{itemize}

\subsubsection*{Llama 3 (Abstract Semantics) vs. Qwen 3 (Direct Semantics)}

This pattern illustrates how Llama 3's secondary vectors often explore a wider and more abstract conceptual space compared to Qwen 3's more direct and action-oriented suggestions.

For the phrase ``Should take,'' Llama 3 suggests abstract responsibilities or concepts one should ``take on,'' while Qwen 3 suggests direct objects or actions.
\begin{itemize}
    \item \textbf{Llama 3 $U_1$}: \texttt{utmost admission inspiration revision discipline quitting responsibility guidance}
    \item \textbf{Qwen 3 $U_1$}: \texttt{refuge aways -away brib} \begin{CJK}{UTF8}{gbsn}半天 (half-day) 午饭 (lunch) \end{CJK} \texttt{away} \begin{CJK}{UTF8}{gbsn}这笔 (this sum)\end{CJK}
\end{itemize}

For ``To imagine,'' Llama 3's vectors include abstract and philosophical concepts to imagine, whereas Qwen 3 focuses on more concrete items like ``scenarios.''
\begin{itemize}
    \item \textbf{Llama 3 $U_1$}: \texttt{reconstruct ethical erect owning peace embodied meanings yourself}
    \item \textbf{Qwen 3 $U_1$}: \texttt{scenarios} \begin{CJK}{UTF8}{gbsn}场景 (scene) \end{CJK} \texttt{scenario} \begin{CJK}{UTF8}{gbsn}也是一种 (is a kind of) \end{CJK} \texttt{oha Scenario worlds Scenario}
\end{itemize}

For ``The discovery,'' Llama 3's vectors describe the impact and nature of a discovery (revolutionary, baffling), while Qwen 3's vectors describe the event of a discovery (a journey, an unintentional bulletin).
\begin{itemize}
    \item \textbf{Llama 3 $U_1$}: \texttt{revolution shed bust of details vind catapult baff}
    \item \textbf{Qwen 3 $U_1$}: \begin{CJK}{UTF8}{gbsn}震惊 (shock) 轶事 (anecdote) 新西 (New West/ New Zealand) 了一个 (a) 之旅 (journey) 无意 (unintentional) 快报 (bulletin) 小镇 (small town)\end{CJK}
\end{itemize}

\subsubsection{Llama 3 (Abstract Semantics) vs. Qwen 3 (Direct Semantics)}
Here, Llama 3's secondary vectors explore broader, more abstract concepts, while Qwen 3's are more concrete and action-oriented.\\

\textbf{14 phrases out of 100} show this strong contrast.

\begin{itemize}
    \item \textbf{Will break:} Llama \texttt{confidentiality, independence}; Qwen \texttt{ties, neck, dance}.
    \item \textbf{Must leave:} Llama \texttt{departing, orientation}; Qwen \texttt{immediately, room}.
    \item \textbf{Should take:} Llama \texttt{admission, inspiration, discipline}; Qwen \texttt{refuge, advantage}.
    \item \textbf{The broken:} Llama \texttt{fragments, promises, torn}; Qwen \texttt{window, clock, vase}.
    \item \textbf{To begin:} Llama \texttt{brainstorm, conceptual}; Qwen \texttt{start, validate}.
    \item \textbf{May open:} Llama \texttt{invitation, plea}; Qwen \texttt{windows, sesame}.
    \item \textbf{Could drive:} Llama \texttt{distracted, fleets, uninsured}; Qwen \texttt{drunk, uphill}.
    \item \textbf{The discovery:} Llama \texttt{revolution, catapult}; Qwen \texttt{journey, bulletin}.
    \item \textbf{To prevent:} Llama \texttt{vulnerability, security}; Qwen \texttt{corrosion, fires}.
    \item \textbf{The solution:} Llama \texttt{vector, lattice, eigen}; Qwen \texttt{set, definition}.
    \item \textbf{To complete:} Llama \texttt{projects, tasks}; Qwen \texttt{orders, assignment}.
    \item \textbf{Were planning:} Llama \texttt{launching, upcoming}; Qwen \texttt{permission, meetings}.
    \item \textbf{The evidence:} Llama \texttt{overwhelmingly, against}; Qwen \texttt{suggests, linking}.
    \item \textbf{To create:} Llama \texttt{customized, empowering}; Qwen \texttt{custom, interactive}.
\end{itemize}

\subsubsection*{Qwen 3's Multilingual Reasoning}


This pattern showcases Qwen 3's unique ability to access a cross-lingual conceptual space, populating its secondary vectors with semantically relevant non-English tokens.

For the phrase ``The fast car,'' Qwen 3's $U_1$ vector includes multiple Chinese words related to speed and motion.
\begin{itemize}
    \item \textbf{Qwen 3 $U_1$}: \texttt{overt} \begin{CJK}{UTF8}{gbsn}的速度 (speed) 运动 (motion) \end{CJK} \texttt{.Speed} \begin{CJK}{UTF8}{gbsn}追赶 (chase) 速度 (speed) \end{CJK} \texttt{riages} \begin{CJK}{UTF8}{gbsn}超越 (surpass)\end{CJK}
\end{itemize}

For ``The fresh bread smelled,'' Qwen 3's $U_2$ vector is a list of Chinese synonyms and related concepts for ``smell'' and ``fragrance.''
\begin{itemize}
    \item \textbf{Qwen 3 $U_2$}: \texttt{smell smells} \begin{CJK}{UTF8}{gbsn}嗅 (sniff/smell) 闻 (smell/hear) 香 (fragrant) 香气 (aroma/fragrance) 香味 (fragrance/scent) \end{CJK} \texttt{smelling}
\end{itemize}

For ``Should help her,'' the $U_1$ vector remarkably contains relevant concepts from multiple languages, including Chinese (career development, alleviate), Russian (cope/handle), and Vietnamese (support/help).
\begin{itemize}
    \item \textbf{Qwen 3 $U_1$}: \begin{CJK}{UTF8}{gbsn}事业发展 (career development) \end{CJK} \foreignlanguage{russian}{справиться (handle/cope)} \foreignlanguage{vietnamese}{hỗ trợ (support/help)} \begin{CJK}{UTF8}{gbsn}缓解 (alleviate) \end{CJK} \texttt{unpack} \begin{CJK}{UTF8}{gbsn}管理工作 (manage work) 过渡 (transition) 学业 (studies)\end{CJK}
\end{itemize}

\textbf{38 phrases out of 100} contain clear examples of multilingual reasoning.

\subsubsection*{Examples of Sub-word Fragments in Qwen 3}

For the phrase ``While walking,'' the second singular vector for the token ``walking'' is almost entirely composed of these fragments, including common suffixes.
\begin{itemize}
    \item \textbf{Vector (Token 1, $U_2$)}: \texttt{\textbf{e} \textbf{ection} \textbf{ing} \textbf{eer} ignKey \textbf{cion} \textbf{ging} \textbf{eed}}
\end{itemize}

For ``To begin,'' the $U_2$ vector includes the common suffixes -ments and -ly, suggesting a mode for building nouns and adverbs.
\begin{itemize}
    \item \textbf{Vector (Token 1, $U_2$)}: \texttt{\textbf{e} \textbf{ments} \textbf{eel} \textbf{hips} \textbf{eed} s \textbf{eve} \textbf{ly}}
\end{itemize}

For ``The deep water,'' the $U_2$ vector for the token ``water'' contains fragments like -ness and -ection.
\begin{itemize}
    \item \textbf{Vector (Token 2, $U_2$)}: \texttt{\textbf{e} y \textbf{eer} \textbf{eel} \textbf{eus} \textbf{ness} \textbf{ection} yth}
\end{itemize}

\subsubsection{Qwen 3's ``Word-Building'' Vector}
This category identifies phrases where a secondary Qwen 3 vector is dominated by sub-word fragments and morphological units (e.g., -ing, -tion, -eer, -ness).\\

\textbf{33 phrases out of 100} clearly display a dedicated morphological vector.

\begin{figure*}[!t]
\centering
\includegraphics[width=0.8\textwidth]{"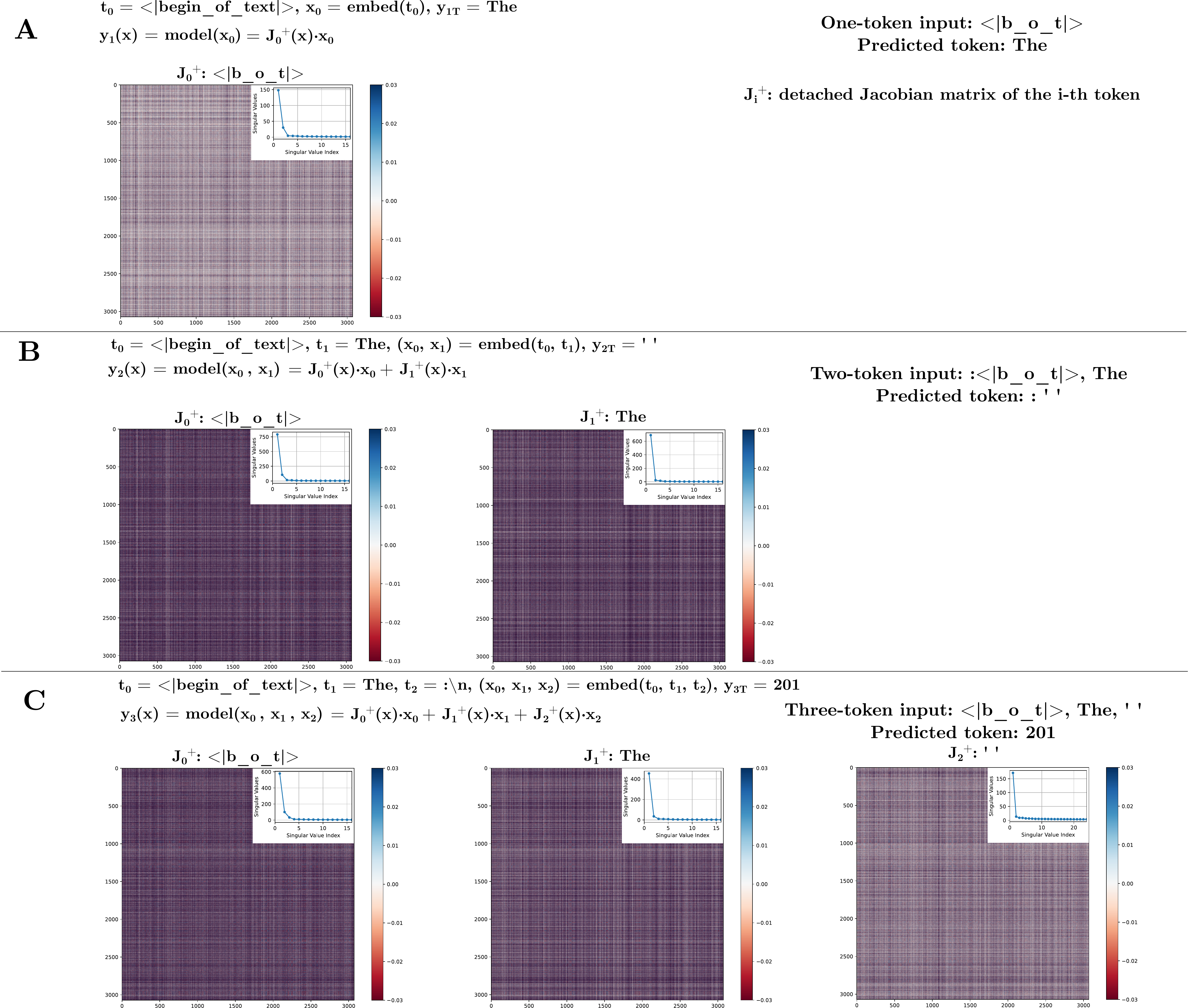"}
\caption{An overview of next-token prediction in the Llama 3.2 3B transformer decoder and decomposition of the predicted embedding vector computation using the detached Jacobian. Generating three tokens with only $<|BoT|>$ as input produces ``The 201''. For each prediction, each input token $\vect{t_{i}}$ is mapped to an embedding vector $\vect{x_{i}}$, and the network generates the embedding of a next token. The phrase turns out to be ``The 2019-2020 season''. The detached Jacobian $\vect{J^+(\vect{x})}$ of the predicted output embedding with respect to the input embeddings is composed of a matrix corresponding to each input vector. Each detached Jacobian matrix $\vect{J^+_{i}}(\vect{x})$ is a function of the entire input sequence but operates only on its corresponding input embedding vector. The matrices tend to be extremely low rank, shown in the inset figures, and the matrix $\vect{J^+_{0}}$ varies across A), B) and C) above because the input sequences differ. Since the detached Jacobian captures the entirety of the model operation in a linear system (numerically, for a given input sequence), tools like the SVD can be used to interpret the model and its sub-components.}\label{fig:jacobian_sum}
\end{figure*}

\begin{figure*}[!t]
\centering
\includegraphics[width=0.99\textwidth]{"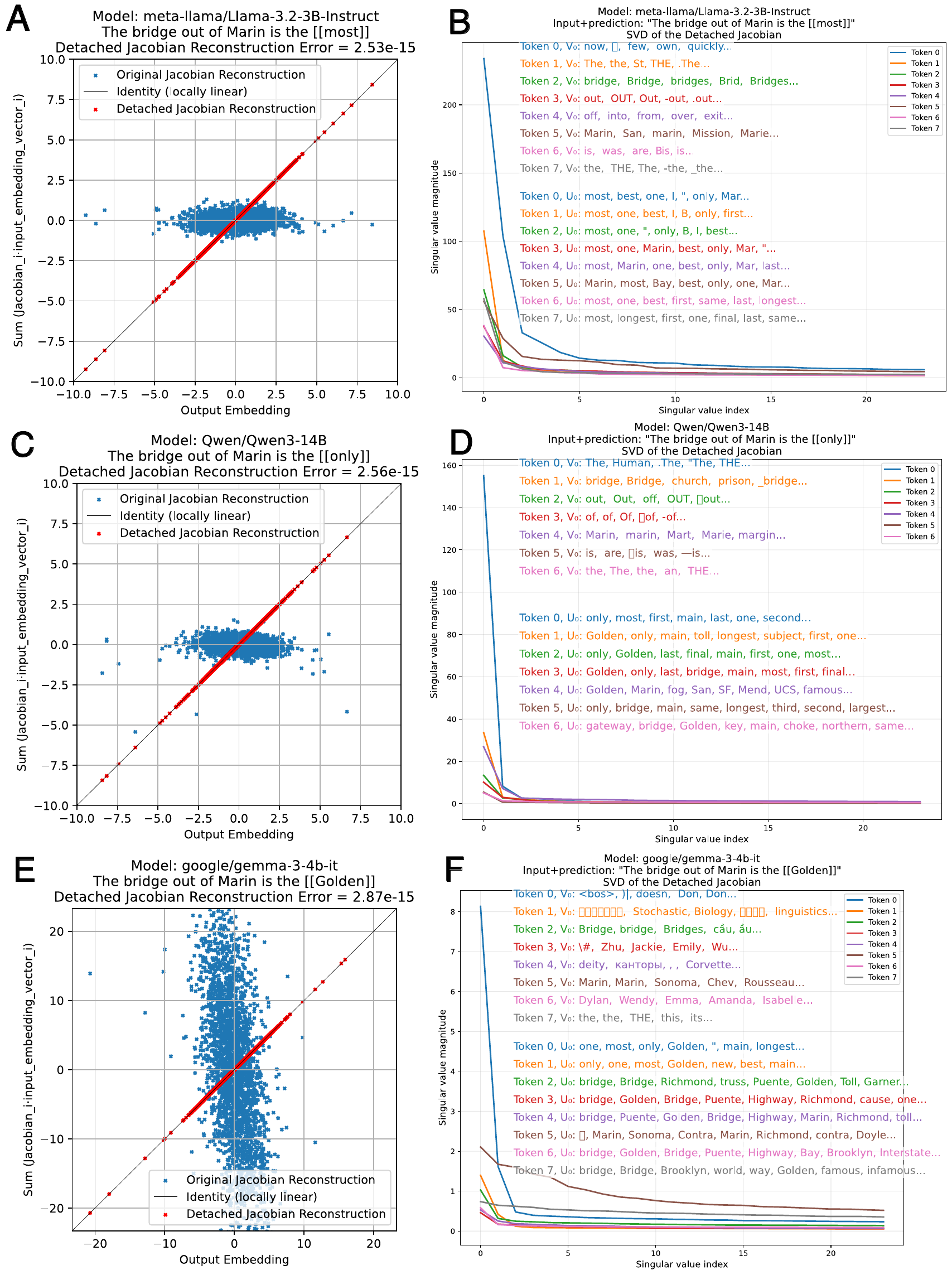"}
\caption{The detached Jacobian reconstruction error and SVD for Llama 3.2 3B, Qwen 3 14B and Gemma 3 4B}\label{fig:equivalent_linear_llama_qwen_gemma}
\end{figure*}

\begin{figure}[!t]
\centering
\includegraphics[width=0.8\columnwidth]{"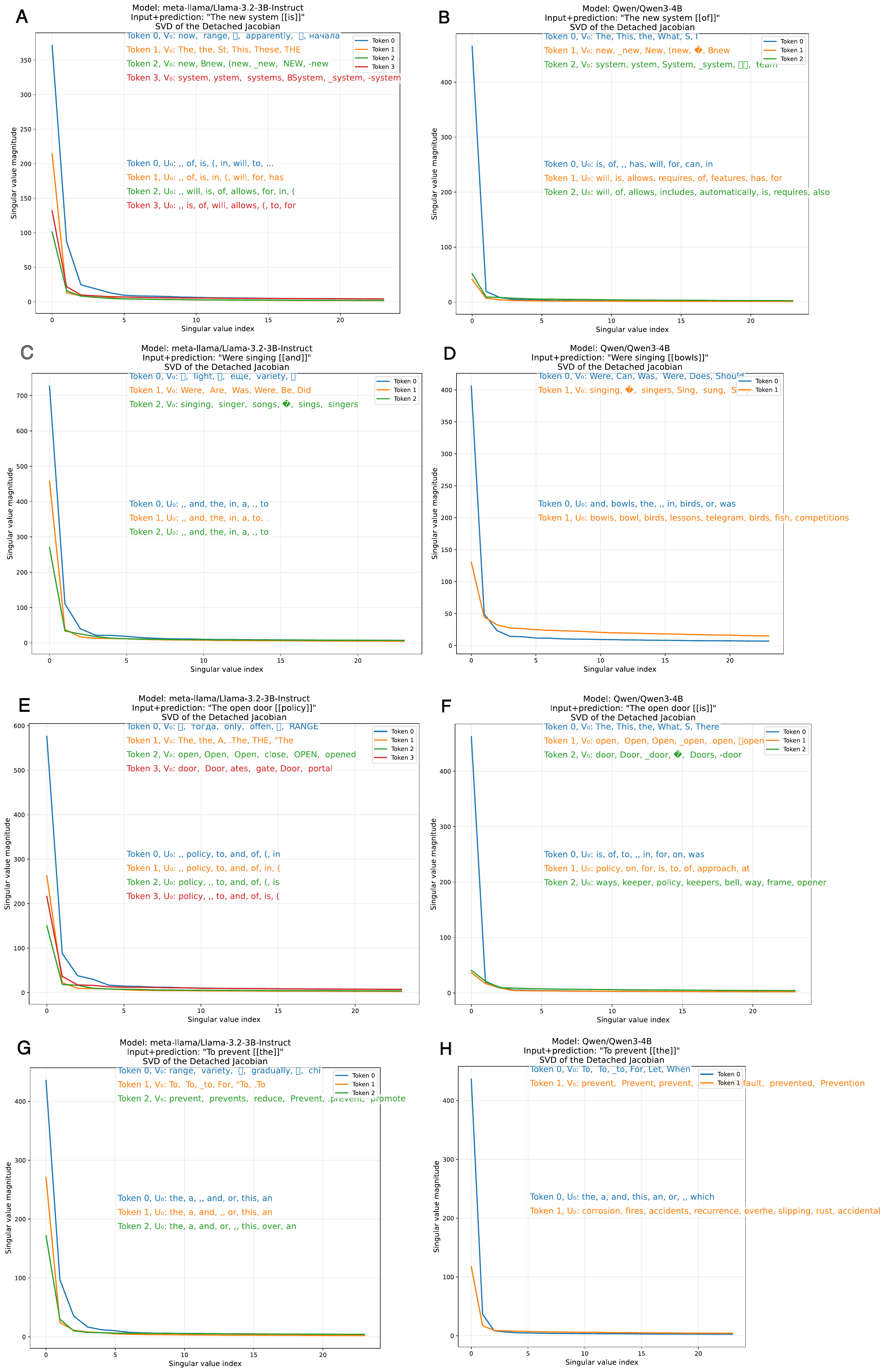"}
\caption{Comparison of detached Jacobians for the same phrase across models.
}\label{fig:across-models}
\end{figure}

\begin{table*}[!ht]
\begin{adjustbox}{width=\textwidth, center}
    \centering \scriptsize
    \begin{tabular}{|l|l|l|l|}
    \hline
    ~ & Input token 0 & Input token 1 & Input token 2 \\ \hline
    \ChangeRT{1.4pt}
    Layer 13\_0 & coli gnu ovny elin ovol ATEG & /Dk MetroFramework olumn   reluct upertino DOT regor & akra izik esteem critical timer noch MUX apest \\ \hline
 Layer 13\_1 & weit LineStyle fonts Ymd ysize rt akra  reverted & ta .unsplash fonts reverted Ymd ograd .gf & Lud QObject darwin adecimal )const angel PushButton usercontent \\ \hline
 Layer 13\_2 & chter i  Burgess  Lud abet  Burke ernal $backslash$Bundle & opia agn  ember  missile  trace osed & card plus imm  cardinal  Spare enz  Eg lex \\ \hline
 \ChangeRT{1.4pt}
    Layer 14\_0 & en  wil  lo ...  764  fa & /Dk HeaderCode [OF ToPoint <typeof spd .liferay NCY & \ weit vc ciler  inx critical akedirs \\ \hline
 Layer 14\_1 & weit  ; dealloc akedirs LineStyle   \hlzero{Bridge} ysize & \ ta defs ones .unsplash arken ; & id rone orm iland  [] .... \\ \hline
 Layer 14\_2 & chter  Burgess  <typeof  Ridge Subsystem  PressEvent & agn   ember  another arend & Kelly  Jar  Cunningham  Jarvis )const  Stadium ortal \\ \hline
 \ChangeRT{1.4pt}
    Layer 15\_0 & wil i  Mage gnu iyah erval  dedicated Mage & skb  abus <typeof ToBounds xcf /Dk   & \hlzero{bridge}   \hlzero{bridges}   \hlzero{Bridge} \   \hlzero{Bridges}   Mood   illin \\ \hline
 Layer 15\_1 & weit \   \hlzero{Bridge} squeeze dealloc   \hlzero{bridges} woke   \hlzero{bridge} & \ illin k .foundation ophe ATEG   \hlzero{bridges} & gc RAD .bunifuFlatButton   Dickinson PushButton NullException   Comet \\ \hline
 Layer 15\_2 & chter <typeof RAD .NULL Subsystem   Burgess ches & arend esser  pair agn kel ered usi ending & Bravo   Cunningham Imm   Brew   losures \\ \hline
 \ChangeRT{1.4pt}
    Layer 16\_0 & i  en  entre lei yre wil ewis iyah & sole /cms \$MESS pdev xcf spd <typeof & \hlzero{bridge}   \hlzero{bridges} \   \hlzero{Bridge} k akra   \hlzero{Bridges}   toll \\ \hline
 Layer 16\_1 & \hlzero{Bridge}   \hlzero{bridge} squeeze \ akra SqlServer weit fabs & \ k 799 izi  Bl akra   .\hlthree{Exit} & gc RAD   Dickinson UpInside   PushButton   clerosis \\ \hline
 Layer 16\_2 & Invalidate RAD <typeof Subsystem /cms   NUITKA saida & arend agn   ag ered  another /out  dipl yll & vere losures   Bravo   Route employment   closure   \hlthree{exit} occo \\ \hline
 \ChangeRT{1.4pt}
    Layer 17\_0 & i  en yre iyah ewis erval   \hlone{only} agnet & sole /cms  gc   \hlzero{Bridge}   \hlzero{bridge} PushButton ToF & \hlzero{bridge}   \hlzero{Bridge}   \hlzero{bridges}   \hlzero{Bridges} k \   choke   Brig \\ \hline
 Layer 17\_1 & \hlzero{Bridge}   Marin   \hlzero{bridge} arLayout Brains   \hlzero{bridges}   choke /connection & k yre .IDENTITY   799 Affected   \hlzero{Bridges} Local & gc PushButton ANNEL   Inspiration   bast   occasion \\ \hline
 Layer 17\_2 & <typeof .scalablytyped   Invalidate   bast   Burke   gc & another isser  hell  to amp arend  new  gender & hosting   closure   Bravo ernal   kond   Hosting   location   Backbone \\ \hline
 \ChangeRT{1.4pt}
    Layer 18\_0 & \hlzero{bridge}   en i yre   end 764  go  San & \hlzero{bridge}   \hlzero{Bridge}   \hlzero{bridges} sole   crossing   brid .\hlzero{bridge}   cause & \hlzero{bridge}   \hlzero{Bridge}   \hlzero{bridges}   \hlzero{Bridges}   brid \hlzero{Bridge} \hlzero{bridge} \\ \hline
 Layer 18\_1 & \hlzero{Bridge}   \hlzero{bridge}   \hlzero{bridges}   Marin \hlzero{Bridge}   \hlzero{Bridges}   brid & \hlzero{Bridges}   \hlzero{bridges}   OUNTRY     .IDENTITY Queries Choices & PushButton gc   Tos ANNEL   Inspiration   Route sole \\ \hline
 Layer 18\_2 & Invalidate .scalablytyped   Burke     Saunders stial   Lair agues & to   another  elsewhere amp bec isser  hell & San ernal   hosting   closure     SF   \^   Bravo \\ \hline
 \ChangeRT{1.4pt}
    Layer 19\_0 & \hlzero{bridge}   SF   \hlthree{exit}   San   \hlone{only}   connecting   Oakland usp & \hlzero{bridge}   \hlzero{Bridge}   \hlzero{bridges}   crossing   toll .\hlzero{bridge}   choke   SF & \hlzero{bridge}   \hlzero{Bridge}   \hlzero{bridges}   \hlzero{Bridges} \hlzero{Bridge} \hlzero{bridge}   brid \\ \hline
 Layer 19\_1 & Marin   \hlzero{Bridge}     \hlzero{bridge}   SF   \hlfour{Golden} \hlzero{Bridge} \hlzero{bridge} & \hlzero{Bridges}   Odds shima iliary ued IID   Oakland .syntax & \hltwo{Highway}   Route     route route ANNEL   Hwy   \hltwo{highway} \\ \hline
 Layer 19\_2 & Fletcher   Los LA   Edgar   Southern   Burke   LA Los & elsewhere     to   new   Oakland   another   hell   progress & SF   San ucker   Oakland   \hlfour{Golden}   Stanford   closure   sf \\ \hline
 \ChangeRT{1.4pt}
    Layer 20\_0 & \hlzero{bridge}   toll   \hlthree{exit}   \hlzero{Bridge}   \hlzero{bridges}   San   SF   Toll & \hlzero{bridge}   \hlzero{Bridge}   toll   \hlzero{bridges}   Toll   tol   crossing .\hlzero{bridge} & toll   \hlzero{bridge}   \hlzero{Bridge}   tol   latest   \hlzero{bridges}   Toll   \hlzero{Bridges} \\ \hline
 Layer 20\_1 & Marin   sf   SF   toll arLayout   Oakland   Berkeley & Odds   Oakland   contra n thing uckles ued istrator & Route     route   \hltwo{Highway} Route odus route   Hwy \\ \hline
 Layer 20\_2 & toFloat PLIED   Southern LA   Los   Fletcher uluk   \hlzero{bridge} & to   elsewhere   new   ella cht   peninsula   syn & SF   Oakland   Stanford anson   San SF .vn   sf \\ \hline
 \ChangeRT{1.4pt}
    Layer 21\_0 & \hlzero{bridge}   toll   \hlzero{bridges}   \hlthree{exit}   San   \hlzero{Bridge}   Toll   SF & \hlzero{bridge}   toll   \hlzero{Bridge}   \hlzero{bridges}   Toll   tol \hlzero{Bridge}   crossing & \hlzero{bridge}   toll   \hlzero{bridges}   \hlzero{Bridge}   \hlzero{Bridges}   tol   Toll \hlzero{Bridge} \\ \hline
 Layer 21\_1 & Marin     SF sf SF   Oakland   Berkeley arLayout & contra   n uckle   Oakland   thing & Route   route .scalablytyped   \hltwo{Highway}   odus Route annel \\ \hline
 Layer 21\_2 & uluk   \hlzero{bridges}   \hlzero{bridge}   \hlzero{Bridge} toFloat   Southern PLIED   Roose & Marin   Oakland   Fog ?type     Berkeley   SF uv & SF   Marin   Oakland SF   San   Stanford   Berkeley   sf \\ \hline
 \ChangeRT{1.4pt}
    Layer 22\_0 & \hlzero{bridge}   toll   San   \hlzero{bridges}   \hlthree{exit}   SF   \hlzero{Bridge}   connecting & \hlzero{bridge}   \hlzero{Bridge}   \hlzero{bridges}   toll   crossing   tol   Toll \hlzero{Bridge} & \hlzero{bridge}   \hlzero{bridges}   toll   \hlzero{Bridge}   \hlzero{Bridges}   tol   crossing \hlzero{Bridge} \\ \hline
 Layer 22\_1 & Marin SF   =\~ inas sf   Berkeley & contra uckle thing 415     .ObjectId n & \hltwo{Highway}   Backbone   Route   \hltwo{highway}   route Route annel .Atomic \\ \hline
 Layer 22\_2 & fabs uluk   \hlzero{bridges} dealloc   Brig GMT   \hlzero{Bridge} anlk & Marin ?type dea   Berkeley   arResult zc occo & SF   Marin SF .sf   sf SF   Salesforce   Berkeley \\ \hline
 \ChangeRT{1.4pt}
    Layer 23\_0 & toll   San   \hlzero{bridge}   \hlthree{exit}   Bay   \hlfour{Golden}   \hlthree{Exit}   connecting & toll   \hlzero{bridge}   \hlfour{Golden}   San   \hlzero{Bridge}   tol   \hlzero{bridges}   Toll & toll   \hlzero{bridge}   \hlzero{bridges}   \hlzero{Bridge}   tol   span   \hlzero{Bridges}   Toll \\ \hline
 Layer 23\_1 & Marin inas =\~   Ukraj   Berkeley & thing riz 415 orte   contra uckle 299 imo & Pacific   Route   coast   \hltwo{Highway}   route .Atomic annel \\ \hline
 Layer 23\_2 & fabs uluk   Brig dealloc pun   Roose   Bud   Cunningham & Marin ?type   Berkeley dea zc ottenham inas aser & Marin   SF   Oakland .sf SF   sf   Berkeley   Bay \\ \hline
 \ChangeRT{1.4pt}
    Layer 24\_0 & \hlzero{bridge}   San   toll   vi   \hlthree{exit}   Bay   SF   \hlfour{Golden} & toll   \hlzero{bridge}   \hlfour{Golden}   San   \hlzero{Bridge}   vi   tol   \hlzero{bridges} & toll   \hlzero{bridge}   span   tol   \hlzero{bridges}   \hlzero{Bridge}     vi \\ \hline
 Layer 24\_1 & Marin sf   /goto   marin     aidu & riz   (   contra rey ued uckle 299   & Pacific   \hltwo{Highway}   Route   Los .Atomic Los   route   \hltwo{highway} \\ \hline
 Layer 24\_2 & uluk fabs dealloc anlk   Brig   Roose simd   Bud & Marin dea ?type zc   chez   Berkeley   app aidu & Marin   SF .sf   sf   Oakland   San SF   Salesforce \\ \hline
 \ChangeRT{1.4pt}
    Layer 25\_0 & toll   \hlzero{bridge}   vi   subject   connecting   \hlthree{exit}   Bay   \hlone{only} & toll   \hlzero{bridge}   \hlfour{Golden}   vi   tol   Toll   \hlzero{Bridge} \hlfour{Golden} & toll   span   \hlzero{bridge}   tol   vi   latest   \hlzero{bridges}   Toll \\ \hline
 Layer 25\_1 & Marin aidu   Skywalker     arm sf & riz iny jev thing 415     contra & Pacific POSIT   Via c   Santa   tracks   rough   Backbone \\ \hline
 Layer 25\_2 & uluk anlk fabs dealloc   Brig   GetEnumerator vka   Roose & Marin   chez dea ?type zc aidu ptr reib & Marin   SF   sf SF .sf   San   Richmond SF \\ \hline
 \ChangeRT{1.4pt}
    Layer 26\_0 & toll   subject   cause   San   \hlfour{Golden}   ge   \hlzero{bridge}   \hlone{only} & \hlfour{Golden}   toll   \hlzero{bridge}   tol   span   cause \hlfour{Golden}   crossing & span   toll   tol   \hlzero{bridge}   spans   \hlfour{Golden}   crossing   vital \\ \hline
 Layer 26\_1 & Marin   aidu .Generated     mainwindow .scalablytyped & 415 arching   iny n   s   contra riz & Pacific   Via   Santa   rough   rou annel Santa   route \\ \hline
 Layer 26\_2 & uluk   Brig anlk   spans .Atomic fabs   Bailey plied & Marin ptr dea ensis aidu   chez lands inas & Marin   sf   SF .sf SF   Richmond   San 415 \\ \hline
 \ChangeRT{1.4pt}
    Layer 27\_0 & subject   \hlone{only}   cause   SF   ge   \hlthree{exit}   connecting   toll & span   toll   \hlfour{Golden}   cause   tol   San   Toll   \hlone{only} & span   toll   tol   spans   vital   Span   symbol   latest \\ \hline
 Layer 27\_1 & Marin     mainwindow /effects   .Generated & arching n pone     s yar & Pacific   Santa .Atomic Santa annel   route   rou POSIT \\ \hline
 Layer 27\_2 & .Atomic uluk   Bailey eview elig stdout   ffset & Marin ensis zc dea   chez lands   agna & Richmond   Marin   SF   sf .sf SF   San SF \\ \hline
 \ChangeRT{1.4pt}
    Layer 28\_0 & Richmond   \hlone{only}   subject   one   last   symbol   toll   I & \hlfour{Golden}   toll   Richmond   tol   span   San   Toll   \hlone{only} & span   toll   symbol   tol   latest   Richmond   \hlone{only}   final \\ \hline
 Layer 28\_1 & P .Generated     /effects & yar pone arching n   s & Pacific .Atomic   Santa   Via   Samuel   twisting     route \\ \hline
 Layer 28\_2 & uluk   Bailey   Ava ffset .Atomic jak   Whip   Santa & Marin odian dea ensis agna agnar zc olin & Richmond   SF   Marin SF   sf .sf SF   San \\ \hline
 \ChangeRT{1.4pt}
    Layer 29\_0 & \hlone{only}   Richmond   final   last   \hlfive{most}   one   subject   symbol & \hlfour{Golden}   \hlfive{most}   tol   toll   final   Richmond   Bay \hlfour{Golden} & tol   last   final   \hlfour{Golden}   symbol   latest   toll   \hlfive{most} \\ \hline
 Layer 29\_1 & \hlfour{Golden} .Iter     .Generated     tslint & pone n arching beiter Ped yar   jev & Via   Santa   Pacific   Samuel   Rim   Corner   winding   coast \\ \hline
 Layer 29\_2 & Marin abra riad seite theless ordion   sonian & Marin olin   marin nov ensis aser agnar   nov & Marin SF   SF .sf   Richmond   sf   San SF \\ \hline
    \end{tabular}
\end{adjustbox}
    \caption{The top three singular vectors of the detached Jacobian for the layer outputs from Llama 3.1 8B for the sequence ``The bridge out of Marin is the'' with the prediction [[Golden]]. Legend: \hlzero{``Bridge''}, \hlone{``only''}, \hltwo{``highway''}, \hlthree{``exit''}, \hlfive{``most''}.}
\label{tab:llama-layers-full}
\end{table*}

\begin{table*}[!ht]
\begin{adjustbox}{width=\textwidth,center}
    \centering \scriptsize
    \begin{tabular}{|l|l|l|l|}
    \hline
 ~ & Input token 0 & Input token 1 & Input token 2 \\ \hline
\ChangeRT{1.4pt}
    Layer 13\_0 & nt  \hlone{only}  the  alors  that ...  and .. & the  that  \hlone{only} nt  and  \hlfive{most}  called  either & probablement  Guad  alcanz  Lans   Yellow   yaitu \\ \hline
 Layer 13\_1 & sh  handful  Shaq fame mselves  pity  mga & Sur GONE ).   Ret  Genau Ret & the  and  that the ... and if nt \\ \hline
 Layer 13\_2 & ively ional )]); & 0 5 1 2 4  mete 3 & Called March Bon Entity Clock Patricia Bin \\ \hline
 \ChangeRT{1.4pt}
    Layer 14\_0 & \hlone{only}  that  the nt ...  one ..  either & \hlone{only}  the  that  one nt  either \hlone{only}  those & vecchio  \hlzero{Bridge}   Guad  probablement  iconic  menambah Night \\ \hline
 Layer 14\_1 & the  1   robot  uvre & Sur GONE MaxLength Toto heus  Seg Novo & the  racist  extremist  that  those  terrorist  scenic  anarch \\ \hline
 Layer 14\_2 &  Tyl & 0 1 5 2 4 3  verwenden 6 & and already after sure Z AL \hlone{only} development \\ \hline
 \ChangeRT{1.4pt}
    Layer 15\_0 & \hlone{only}  called  one   probably  either  the  very & \hlone{only}  the  one  either  that  called  probably  transportation & iconic  \hlzero{Bridge}  \hlzero{bridge}  IQR Nope  puente  \hlzero{bridges}  \hlzero{Bridges} \\ \hline
 Layer 15\_1 & feminism 1 robot   imperialism  ems  polticas & tzw Seg  tzw North Sur Secondo  aman  atraves & one  extremist  racist  the  camping  terrorist  supposed  military \\ \hline
 Layer 15\_2 & )  ). & 0 1 5 2  verwenden 3 4 6 &  called   Sept   (   Grand \\ \hline
 \ChangeRT{1.4pt}
    Layer 16\_0 &  \hlone{only}   one  very  first  the   route & \hlone{only}   the  one  first  three  two  very & iconic  \hlzero{Bridge}  \hlzero{bridge}  Centennial  Greater  \hlfour{Golden}  \hlzero{bridges}  Iconic \\ \hline
 Layer 16\_1 & tzw North  bernama Nec  tangent Bitter Seg getAvg & North  North  tzw  tzw  largest  lagoon  sogenannten  shortcut & coastal  scenic  not  military  one  likely  very  location \\ \hline
 Layer 16\_2 & Arc   Approx & 0 5 1 2 3  Provides 6 4 & turbo Hydro  Turbo Geo Mapa  northward  blasted  north \\ \hline
 \ChangeRT{1.4pt}
    Layer 17\_0 & \hlone{only}   first  \hlfive{most}  one  very  two  the & \hlone{only}   one  first  the  two  \hlfive{most}  very & iconic  legendary  famed  famous  Greater  namesake  infamous  Centennial \\ \hline
 Layer 17\_1 & hacerlo  tangent  tzw  bernama  north  loophole & North  north North  tzw  tzw  sogenannte  sogenannten  behem & \hlone{only}  very  not  scenic  one  likely  military  extremely \\ \hline
 Layer 17\_2 & ).  Paths Tub .). ) Endpoints Heap & 0 1 5 2 3  Provides 4  verwenden & Turbo  TOC    Pipeline  Aviation  City  Route  Water \\ \hline
 \ChangeRT{1.4pt}
    Layer 18\_0 & \hlone{only}   first  \hlfive{most}  one  very  two  the & \hlone{only}   first  one  \hlfive{most}  two  the  that & \hlzero{bridge}  \hlzero{Bridge}  \hlzero{bridges}   iconic  crossing  \hlzero{Bridges} \\ \hline
 Layer 18\_1 & loophole Nec Locator FBSDKAccessToken  peanut  sebaik & behem  giant  loophole  lagoon  swamp MaxLength Nec Locator & military  \hlone{only}  coastal  area  likely  location  beaches  areas \\ \hline
 Layer 18\_2 & ).   .).   Paths & 0  Design 1  Style 3  Styling  Provides & Water  Pipeline  Watercolor  Pipelines  Fountain  Beacon  Marathon  Balloon \\ \hline
 \ChangeRT{1.4pt}
    Layer 19\_0 & \hlone{only}  first  one  \hlfive{most}  two  \hltwo{highway}  the & \hlone{only}  one  first  two  that  \hlfive{most}  second & \hlzero{Bridge}  \hlzero{bridge}  \hlzero{bridges}  \hlzero{Bridges}  Crossing \hlzero{Bridge}  \hlfour{Golden}  crossing \\ \hline
 Layer 19\_1 & Ring  Coc   Road  reverse  Rd  Beacon & \hlzero{Bridge}  Ring  Coc  Road  notorious  iconic  behem  Reverse & coastal  military  area  location  areas  \hlfive{most}  beach  beaches \\ \hline
 Layer 19\_2 & vreau rupani  nggak  inclusin  advogado  comprens   emphas & Crossing  Compre  Laufe  Steel   Indem  Cht  \hlzero{Bridge} & Aviation  Outreach  Pipelines  Turbo  Wastewater  Pipeline  Brewing  Beacon \\ \hline
 \ChangeRT{1.4pt}
    Layer 20\_0 &  one  \hlone{only}  \hlfive{most}   first   second  very &  one  \hlone{only}   \hlfive{most}  first  second  best   & \hlzero{Bridge}  \hlzero{bridge}  \hlzero{bridges}  Crossing  \hlzero{Bridges}  crossing  bridging  iconic \\ \hline
 Layer 20\_1 & Mun Mun Har  Tak  Coc  Trinity  Beacon & Har  Har  Est  Tak  Mega   Rainbow  Coc & area  one  areas  \hlfive{most}  location  \hlone{only}  coastal  military \\ \hline
 Layer 20\_2 & Zap Typical  Ric Tub +)\$ Ridge & \hlzero{Bridge}  Crossing  \hlzero{Bridges} Crossing  \hlzero{Bridge}   Guad & teapot  minus  Vectors  Spiral  spiral  turbo   spirals \\ \hline
 \ChangeRT{1.4pt}
    Layer 21\_0 & \hlzero{bridge}  \hlone{only}  one  \hlfive{most}  \hltwo{highway}   first  best & \hlzero{bridge}  \hlone{only}  one  \hlfive{most}  first   \hltwo{highway}  best & \hlzero{Bridge}  \hlzero{bridge}  \hlzero{bridges} \hlzero{Bridge}  \hlzero{Bridges}  Crossing  crossing  bridging \\ \hline
 Layer 21\_1 & \hlzero{Bridge} \hlzero{Bridge}   \hlzero{bridge}  \hlzero{bridges} zungen  Tol  Tak & \hlzero{Bridge}  \hlzero{bridges} \hlzero{Bridge}  \hlzero{bridge}   \hlzero{Bridges}  bridging  Tol & \hlfive{most}  coastal  area  areas  maritime  one  closest  fastest \\ \hline
 Layer 21\_2 & vreau  gobierno  totalit   comprens  nggak  lackluster  ejrcito & \hlzero{Bridge}  Crossing  \hlzero{Bridges}  Ponte Crossing \hlzero{Bridge} & mansion  Architectural  Sculpture  Basilica  monumento Monument Museum  edifice \\ \hline
 \ChangeRT{1.4pt}
    Layer 22\_0 & \hlone{only}  \hlzero{bridge}  one  \hlfive{most}  first  \hltwo{highway}  new & \hlone{only}  \hlzero{bridge}  one  \hlfive{most}  first  new   \hltwo{highway} & \hlzero{Bridge}  \hlzero{bridge}  \hlzero{bridges}  \hlzero{Bridges} \hlzero{Bridge}   bridging  Ponte \\ \hline
 Layer 22\_1 & Namara  Puente  chercher zungen  McCullough  klnb  Wheeler & \hlzero{Bridge}  Puente  puente  \hlzero{bridges}   Tol \hlzero{Bridge} & \hltwo{highway}  road  route  roads  fastest  \hlfive{most}  coastal  coast \\ \hline
 Layer 22\_2 & ERISA  vreau  lackluster   gacche  pabbaj  ceremonia  prosa  i & \hlzero{Bridge}     \hlzero{Bridges} \hlzero{Bridge}  Geral   Design & routes  highways  roads routes  route  ferries  freeway  expressway \\ \hline
 \ChangeRT{1.4pt}
    Layer 23\_0 & \hlone{only}  \hlzero{bridge}  California  \hlfive{most}  one  \hltwo{highway}  freeway  \hlzero{Bridge} & \hlone{only}  \hlzero{bridge}  California  one  \hlfive{most}  \hltwo{highway}  first  new & \hlzero{bridge}  \hlzero{Bridge}  \hlzero{bridges}  \hlzero{Bridges} \hlzero{Bridge}  bridging \\ \hline
 Layer 23\_1 & zungen  Langer <unused58> qualiter loadNpmTasks  menghilangkan & \hlzero{Bridge}  \hlzero{Bridges}  \hlzero{bridges} <unused58>  puente \hlzero{Bridge}  Langer & California  freeway  coastal  Pacific  trailhead  fastest  Californian  route \\ \hline
 Layer 23\_2 & Sonoma  Marin  Napa Marin  Esprito  Medford  California & \hlzero{Bridge}   \hlzero{Bridge}  \hlzero{Bridges}    \hlzero{bridge}  \hlzero{bridges} & Sonoma  Californians California  Californian   Valle  Monterey  Yosemite \\ \hline
 \ChangeRT{1.4pt}
    Layer 24\_0 & California  \hlone{only}  \hlzero{bridge}  freeway  \hltwo{highway}  \hltwo{Highway}  Pacific  \hlzero{Bridge} & California  \hlone{only}  \hlzero{bridge}  freeway  \hltwo{highway}  \hltwo{Highway}  \hlzero{Bridge}  one & \hlzero{bridge}  \hlzero{Bridge}  \hlzero{bridges} \hlzero{Bridge}  \hlzero{Bridges} \hlzero{bridge}  bridging  puente \\ \hline
 Layer 24\_1 & <unused58> pored  !  \hlzero{Bridges}  Langer & \hlzero{Bridges}  \hlzero{bridges}  \hlzero{Bridge} \hlzero{Bridge}  puente <unused58>  bridging & freeway  \hltwo{highway}  route  trailhead  fastest  roads  highways  pathway \\ \hline
 Layer 24\_2 & Sonoma  Swiss Essex Marin  Esprito  Marin  Medford & \hlzero{Bridges}  \hlzero{Bridge}   \hlzero{Bridge}   \hlzero{bridges} \hlzero{bridges} & routes  Routes Route routes  Route  route route  Routing \\ \hline
 \ChangeRT{1.4pt}
    Layer 25\_0 & \hlzero{bridge}  \hlzero{Bridge}  \hlzero{bridges}  California  \hlone{only} \hlzero{bridge}  \hltwo{Highway}  \hltwo{highway} & \hlzero{bridge}  \hlzero{Bridge}  \hlzero{bridges}  California  \hlone{only} \hlzero{bridge}  ferry  \hltwo{Highway} & \hlzero{bridge}  \hlzero{Bridge}  \hlzero{bridges} \hlzero{Bridge}  \hlzero{Bridges} \hlzero{bridge}  bridging \hlzero{bridges} \\ \hline
 Layer 25\_1 & \hlzero{Bridge}  \hlzero{Bridges}  \hlzero{Bridge}  \hlzero{bridges}  \hlzero{bridge}  bridging  puente & \hlzero{bridges} \hlzero{Bridge}  \hlzero{Bridges}  \hlzero{Bridge}  \hlzero{bridge}  bridging  puente & \hltwo{highway}  route  trailhead  freeway  roads  road  \hltwo{Highway}  trail \\ \hline
 Layer 25\_2 & Marin  Marin  Burmese   SF  SF  Sonoma  Genova & \hlzero{Bridges}  \hlzero{Bridge} \hlzero{Bridge}  \hlzero{bridges}   \hlzero{bridges} & Omaha  Wichita  Milwaukee  Houston  Memphis  Chicago  Nebraska  Detroit \\ \hline
 \ChangeRT{1.4pt}
    Layer 26\_0 & \hlzero{bridge}  \hlzero{Bridge}  \hlzero{bridges}  California  \hlone{only}  San  \hltwo{Highway}  \hlfive{most} & \hlzero{bridge}  \hlzero{Bridge}  \hlzero{bridges}  California  \hlone{only}  San  \hltwo{Highway}  \hlfive{most} & \hlzero{bridge}  \hlzero{Bridge}  \hlzero{bridges} \hlzero{Bridge}  \hlzero{Bridges} \hlzero{bridge}  bridging \hlzero{bridges} \\ \hline
 Layer 26\_1 & \hlzero{Bridges}  \hlzero{bridges}  \hlzero{Bridge}  puente  \hlzero{bridge}  bridging & \hlzero{bridges}  \hlzero{Bridges}  \hlzero{Bridge}  puente  \hlzero{bridge}  \hlzero{Bridge}  bridging & route  \hltwo{highway}  trailhead  freeway  pathway  \hltwo{Highway}  fastest  road \\ \hline
 Layer 26\_2 & Marin  Marin  sf SF  Burmese  SF  Sonoma & \hlzero{Bridge} \hlzero{Bridge}  \hlzero{Bridges} & Utah  Angkor  Boise  Nebraska  Alabama  Omaha  Mormon \\ \hline
 \ChangeRT{1.4pt}
    Layer 27\_0 & \hlzero{bridge}  \hlzero{bridges}  \hlzero{Bridge}  California  \hlone{only}  \hlfive{most}  \hltwo{Highway}  \hltwo{highway} & \hlzero{bridge}  \hlzero{bridges}  \hlzero{Bridge}  California  \hlone{only}  \hlfive{most} \hlzero{bridge}  \hltwo{Highway} & \hlzero{bridge}  \hlzero{Bridge}  \hlzero{bridges} \hlzero{Bridge}  \hlzero{Bridges} \hlzero{bridge}  bridging  puente \\ \hline
 Layer 27\_1 & \hlzero{Bridges}   puente  \hlzero{bridges} <unused25> \hlzero{Bridge} & \hlzero{Bridges}  \hlzero{bridges}   puente  bridging \hlzero{Bridge}  \hlzero{bridge}  \hlzero{Bridge} & route  \hltwo{highway}  trailhead  pathway  freeway  \hlfive{most}  trail  path \\ \hline
 Layer 27\_2 & Marin  Marin  Sonoma  Burmese   sf & Struct  Structural Structural Struct & Utah  Mormon  Boise  Angkor  Alabama  Cebu  Birmingham  Nebraska \\ \hline
 \ChangeRT{1.4pt}
    Layer 28\_0 & \hlzero{bridge}  \hlzero{Bridge}  \hlzero{bridges}  \hlone{only}  \hlfive{most}  California  San  \hltwo{Highway} & \hlzero{bridge}  \hlzero{Bridge}  \hlzero{bridges}  \hlone{only}  \hlfive{most}  California  San  Pacific & \hlzero{bridge}  \hlzero{Bridge}  \hlzero{bridges} \hlzero{Bridge} \hlzero{bridge}  \hlzero{Bridges}  bridging  puente \\ \hline
 Layer 28\_1 & wachung oksatta athermy  puente & puente   \hlzero{Bridges}  \hlzero{bridges}  bridging <unused58> athermy & \hltwo{highway}  route  pathway  freeway  \hltwo{Highway}  \hlfive{most}  path  gateway \\ \hline
 Layer 28\_2 & Marin  Marin  Sonoma   marin  marin kafka & Struct Struct & Sonoma     ruari  yaml \\ \hline
 \ChangeRT{1.4pt}
    Layer 29\_0 & \hlone{only}  \hlfive{most}  \hlzero{bridge}  one  \hlfour{Golden}  California  longest  largest & \hlone{only}  \hlfive{most}  \hlzero{bridge}  one  \hlfour{Golden}  longest  California  largest & \hlzero{bridge}  \hlzero{Bridge}  \hlzero{bridges} \hlzero{Bridge} \hlzero{bridge}  puente  bridging \\ \hline
 Layer 29\_1 & wachung azitt patx athermy orragie & puente  bridging  \hlzero{bridges}   \hlzero{Bridges}   \hlzero{Bridge} & route  \hltwo{highway}  trail  path  gateway  pathway  trailhead  \hlfive{most} \\ \hline
 Layer 29\_2 & Marin  Marin   Sonoma  kafka   Ukraj & <unused2146> struct Structure  Struct  Structural  Structure & Snapshot   nt   Sonoma \\ \hline
 \ChangeRT{1.4pt}
    Layer 30\_0 & \hlfive{most}  one  \hlone{only}  \hlzero{bridge}  California  new  \hlfour{Golden} & \hlfive{most}  \hlone{only}  one  California  new  \hlzero{bridge}  \hlfour{Golden}  longest & \hlzero{bridge}  \hlzero{Bridge}  \hlzero{bridges} \hlzero{Bridge}   puente \hlzero{bridge}  toll \\ \hline
 Layer 30\_1 & wachung arakatuh  Choibalsan athermy & bridging  puente   \hlzero{bridges}   getTransforms & route  \hltwo{highway}  trail  trailhead  path  pathway  trails  gateway \\ \hline
 Layer 30\_2 & Marin   Marin  marin   Sonoma & struct   Struct  Structure & prescribe  Snapshot nt \\ \hline
 \ChangeRT{1.4pt}
    Layer 31\_0 & \hlfive{most}  \hlone{only}   one  \hlzero{bridge}  new  main  \hlfour{Golden} & \hlone{only}  \hlfive{most}  one   new  main  \hlfour{Golden}  \hlzero{bridge} & \hlzero{bridge}  \hlzero{Bridge}  \hlzero{bridges} \hlzero{Bridge} \hlzero{bridge}  toll  puente  Toll \\ \hline
 Layer 31\_1 & bottlene  Comunic azitt lytres  qttr & puente   \hlzero{bridges}   lytres & route  trail  \hltwo{highway}  path  pathway  trails  trailhead  \hltwo{Highway} \\ \hline
 Layer 31\_2 & Marin  Marin    kuk & structures & Alabama  Idaho  Kansas  Angkor  Oklahoma  Nebraska  dunes  fuselage \\ \hline
    \end{tabular}
\end{adjustbox}
\caption{The top three singular vectors of the detached Jacobian for the layer outputs from Gemma 3 4B for the sequence ``The bridge out of Marin is the'' with the prediction [[Golden]]. Legend: \hlzero{``Bridge''}, \hlone{``only''}, \hltwo{``highway''}, \hlthree{``exit''}, \hlfive{``most''}.}
\label{tab:gemma-layers-full}
\end{table*}

\begin{table*}[!ht]
\begin{adjustbox}{width=\textwidth, center}
    \centering \scriptsize
    \begin{tabular}{|l|l|l|l|}
    \hline
~ & Input token 0 & Input token 1 & Input token 2 \\ \hline
    Layer 20\_0 & TRY  NORMAL &  massage & akedown eway slow    congest  nodeId \\ \hline
 Layer 20\_1 & AUSE   nrw  bbw  metaphor .listFiles  stret tgt & overlay  extracts  Liter & villa fashion  getattr  depress   bias \\ \hline
 Layer 20\_2 & ade flutter  Fil mon imm   and ren & lyr bounding while & Entities campaign  EventBus   .FILL \\ \hline
 \ChangeRT{1.4pt}
    Layer 21\_0 & TRY    REGARD    dT & REGARD    massage & eway slow  exiting  outbound    fastest  tight \\ \hline
 Layer 21\_1 & @end IGHL ocos UAGE crt & overlay  substr    tag  adorn  bestowed & Managed  meds Choices  TORT  Madness machine  Spare \\ \hline
 Layer 21\_2 & Terr  tag iers imm  Fil ues  Mal & itol  Tomorrow  goodbye stash  calar  lyr  syrup & HTTPS  reinterpret  UTF  REFER  JSON   Netflix \\ \hline
 \ChangeRT{1.4pt}
    Layer 22\_0 & tweaking CONSTANTS & vacc      getch Period & first  hardest  fastest  exiting  ramp \\ \hline
 Layer 22\_1 & metaphor   unc  DERP OBJC  stret .wp  ISP & substr MBOL    \hlzero{bridge} & hurry HIP  opi  Rockets   TORT \\ \hline
 Layer 22\_2 & alk ole ool ros angan icon  vn & antics  ikerrocking & backstory  weblog  SVG  JSON   INCIDENT \\ \hline
 \ChangeRT{1.4pt}
    Layer 23\_0 & salopes   CONSTANTS  getch   Uncomment massage TRY & metaphor  \hlzero{bridge}   largest  easiest  \hlone{only}  first  centerpiece & first   hardest  unc  fastest   \hltwo{highway}  bottleneck \\ \hline
 Layer 23\_1 & metaphor    unc   Derne  makeshift OBJC & \hlzero{bridge}  \hlzero{Bridge} .\hlzero{bridge}  \hlzero{bridge} & scenes WithError   opi Timing   presets Entering \\ \hline
 Layer 23\_2 & ros  lovers flutter & antics   jams & weblog  COMPONENT    annot   metaphor \\ \hline
 \ChangeRT{1.4pt}
    Layer 24\_0 & first  third  last  \hlfive{most}  largest  fourth  culmination & metaphor  largest  centerpiece  easiest  first  hardest  \hlzero{bridge}  gateway & hardest  first  easiest  fastest  \hlfive{most}   ones  same \\ \hline
 Layer 24\_1 & metaphor  makeshift    .wp REAK & brid \hlzero{bridge}  \hlzero{bridges}  \hlzero{bridge}  \hlzero{Bridge} & scenes LocalStorage   WithError \\ \hline
 Layer 24\_2 & flutter rosLingu & ~ & phenomena   puzz  annot    metaphor \\ \hline
 \ChangeRT{1.4pt}
    Layer 25\_0 & largest  \hlfive{most}  first  longest  latest  fastest  last  third & \hlzero{bridge}  \hlzero{bridges}     \hlzero{Bridge}  gateway & hardest  ones  \hlthree{exit}  easiest  first  \hlfive{most}  fastest  \hltwo{highway} \\ \hline
 Layer 25\_1 & \hlzero{bridge}    \hlzero{bridges}   \hlzero{Bridge}  \hlzero{Bridges}  brid & \hlzero{bridges}  \hlzero{bridge}  \hlzero{Bridge} \hlzero{bridge}  \hlzero{Bridge} & (\hlthree{exit} \hlthree{exit}  exits  eternity .\hlthree{exit} \\ \hline
 Layer 25\_2 & \hlzero{bridges}  \hlzero{bridge}  \hlzero{Bridge}  \hlzero{bridge}   parliament & \hlthree{Exit}  \hlthree{exit}  jams & INCIDENT     symbolism \\ \hline
 \ChangeRT{1.4pt}
    Layer 26\_0 & first  \hlfive{most}  largest  last  longest  latest  gateway  \hlone{only} & \hlzero{bridge}  \hlzero{bridges}    metaphor  gateway  connecting & \hltwo{highway}  first  \hlthree{exit}  ones  last   hardest  roads \\ \hline
 Layer 26\_1 & \hlzero{bridge}  \hlzero{bridges}    metaphor  \hlzero{Bridges}  \hlzero{Bridge} & \hlzero{bridges}   \hlzero{bridge}  structures   brid \hlzero{bridge} & .charset    jams Margins \\ \hline
 Layer 26\_2 & parliament  structures   \hlzero{bridges}  Parliament    \hlzero{bridge} & \hlthree{Exit}   \hlthree{exit}  choke  \hlthree{Exit}  panicked & symbolism    metaphor \\ \hline
 \ChangeRT{1.4pt}
    Layer 27\_0 & first  last  largest  \hlzero{bridge}  longest  \hlfive{most}  oldest  latest & \hlzero{bridge}  \hlzero{bridges}     \hlzero{Bridge}   \hlzero{Bridges} & last  first  \hlthree{exit}  \hltwo{highway}  bottleneck  next  road  choke \\ \hline
 Layer 27\_1 & \hlzero{bridge}   \hlzero{bridges}    \hlzero{Bridge}  \hlzero{Bridges} & \hlzero{bridges}   \hlzero{bridge}  \hlzero{Bridge} \hlzero{bridge}  brid \hlzero{Bridge} & EXIT \hlthree{exit}  exits   (\hlthree{exit} \\ \hline
 Layer 27\_2 & \hlzero{bridge} \hlzero{bridge}  \hlzero{bridges}   \hlzero{Bridge}    structures & \hlthree{Exit} \hlthree{exit}  \hlthree{Exit} 	\hlthree{exit} .\hlthree{exit} & incident  EXTRA   incidents \\ \hline
 \ChangeRT{1.4pt}
    Layer 28\_0 & \hlzero{bridge}  longest  largest  first  busiest  last  oldest  \hlfive{most} & \hlzero{bridge}  \hlzero{bridges}     \hlzero{Bridge}  \hlzero{Bridge} & \hltwo{highway}  \hlthree{exit}  bottleneck  highways  \hltwo{Highway}  last  road  exits \\ \hline
 Layer 28\_1 & \hlzero{bridge}  \hlzero{bridges}     \hlzero{Bridge}  \hlzero{Bridge} & \hltwo{highway}  highways   coast  freeway  roads  road  route & \hlthree{exit}  exits    EXIT   \hlthree{exit} \hlthree{Exit} \\ \hline
 Layer 28\_2 & \hlzero{bridge} \hlzero{bridge}  \hlzero{bridges}    \hlzero{Bridge}  brid & \hlthree{Exit}  \hlthree{exit}  \hlthree{Exit}  \hlthree{exit}  \hlthree{exit} 	\hlthree{exit} & Saddam  Mosul  Kuwait     incident  metaphor \\ \hline
 \ChangeRT{1.4pt}
    Layer 29\_0 & \hlzero{bridge}  \hlone{only}  fourth  last  third  longest  fifth  \hlfive{most} & \hlzero{bridge}  \hlzero{bridges}    \hlzero{Bridge}    \hlzero{Bridges} & \hlone{only}  last  first  \hltwo{highway}  third  highways  \hlthree{exit}  fourth \\ \hline
 Layer 29\_1 & \hlzero{bridge}  \hlzero{bridges}     \hlzero{Bridge}  \hlzero{Bridges} & coast  \hltwo{highway}  road  driveway  coastline  roads  highways  freeway & exits \hlthree{exit}    EXIT \\ \hline
 Layer 29\_2 & \hlzero{bridge}  \hlzero{bridges} \hlzero{bridge}   structures  brid   structure & \hlthree{Exit} \hlthree{exit}   \hltwo{Highway}  \hlthree{Exit} & Saddam    Mosul  Elvis  metaphor  incident \\ \hline
 \ChangeRT{1.4pt}
    Layer 30\_0 & \hlzero{bridge}  \hlfive{most}  longest  fourth  third  last  \hlone{only}  fifth & \hlzero{bridge}  \hlzero{bridges}    \hlzero{Bridge}   \hlzero{Bridge} & \hltwo{highway}  \hlone{only}  \hlzero{bridge}  last  first  \hltwo{Highway}  road  highways \\ \hline
 Layer 30\_1 & \hlzero{bridge}   \hlzero{bridges}    \hlzero{Bridges}   \hlzero{Bridge} & coast  freeway   \hltwo{highway}  coastline  road  roads  highways & \hlzero{bridge}   \hlzero{Bridge}  \hlzero{bridges} \hlzero{bridge}   brid \\ \hline
 Layer 30\_2 & \hlzero{bridge}  structure  structures  \hlzero{bridges} \hlzero{bridge}   brid & sail  seab    sailing  Bermuda  ship & Memphis  Kuwait  Jordan   Saddam   Iowa \\ \hline
 \ChangeRT{1.4pt}
    Layer 31\_0 & \hlzero{bridge}  \hlfive{most}  \hlone{only}  last  longest  first  third  largest & \hlzero{bridge}   \hlzero{bridges} \hlzero{Bridge}   \hlzero{Bridge} & \hlone{only}  last  \hltwo{highway}  first  \hlzero{bridge}  \hlthree{exit}  \hltwo{Highway}  \hlfive{most} \\ \hline
 Layer 31\_1 & coast  airlines Interior  airline  interior  Lua  Speedway & coast  coastline  coastal   Coast  route  beach  Coastal & \hlzero{bridge}   \hlzero{Bridge}  \hlzero{bridges}  \hlzero{bridge}  underwater  brid \\ \hline
 Layer 31\_2 & \hlzero{bridge}   \hlzero{bridges} \hlzero{bridge}  brid  \hlzero{Bridge}   structure & ship    sail   sailing  dock  seab & Jordan  Memphis  Kuwait     Mississippi \\ \hline
 \ChangeRT{1.4pt}
    Layer 32\_0 & \hlzero{bridge}  \hlfive{most}  \hlone{only}  first  last  longest  third  largest & \hlzero{bridge}   \hlzero{Bridge}  \hlzero{bridges}  \hlzero{Bridge}  \hlzero{bridge} & \hlone{only}  last  first  \hltwo{highway}  \hlfive{most}  main  route  \hlthree{exit} \\ \hline
 Layer 32\_1 & interior  airline  steam  airlines Trail  breed   vacuum & coast  coastal  coastline  route   Coast  Route  beach & \hlzero{bridge}  span  underwater  connecting  deck  public   member \\ \hline
 Layer 32\_2 & \hlzero{bridge} \hlzero{bridge}   \hlzero{bridges}   \hlzero{Bridge}  brid \hlzero{Bridge} & ship  sail    dock  sailing   seab & Kuwait    Jordan  Memphis  Edmonton  Mississippi  Nile \\ \hline
 \ChangeRT{1.4pt}
    Layer 33\_0 & \hlone{only}  first  last  \hlfive{most}  third  main  second  subject & \hlzero{bridge}  \hlzero{Bridge}   \hlzero{bridges} \hlzero{Bridge}   \hlone{only} & \hlone{only}  last  first  key  main  same  \hlfive{most}  \hlthree{exit} \\ \hline
 Layer 33\_1 & planet   interior  cabin  floors  roots & coast  coastline  coastal   Coast  route  beach  Coastal & span  public  member  library  platform  floating  intervening  deck \\ \hline
 Layer 33\_2 & \hlzero{bridge} \hlzero{bridge}   structure  \hlzero{bridges}  brid  \hlzero{Bridge} & ship   orbit   aircraft  sail  vessel & Kuwait    Nile  Edmonton  Saskatchewan  Tulsa \\ \hline

\end{tabular}
\end{adjustbox}    
\caption{The top three singular vectors of the detached Jacobian for the layer outputs from Qwen 3 14B for the sequence ``The bridge out of Marin is the'' with the prediction [[only]]. Legend: \hlzero{``Bridge''}, \hlone{``only''}, \hltwo{``highway''}, \hlthree{``exit''}, \hlfive{``most''}.}
\label{tab:qwen-layers-full}
\end{table*}

\end{document}